\documentclass[twoside,11pt]{article}

\usepackage[accepted, specialissue]{melba}

\usepackage{verbatim}
\usepackage{array, multirow, caption, booktabs}
\usepackage{subcaption}
\usepackage[dvipsnames]{xcolor}

\usepackage{mwe} 

\usepackage{amsmath,amsfonts}

\melbaid{2023:005}  
\melbaauthors{Bhat, Pluim, Viergever, and Kuijf}  
\volume{2}
\firstpageno{151}  
\melbayear{2023}  
\datesubmitted{12/2022}  
\datepublished{04/2023}  

\melbaspecialissue{MICCAI 2022 UNSURE Workshop}
\melbaspecialissueeditors{Christian Baumgartner, Adrian Dalca, Koen Van Leemput, Raghav Mehta,\\ Chen Qin, Carole Sudre, Ryutaro Tanno, William (Sandy) Wells}

\ShortHeadings{Effect of latent space distribution on the segmentation of images with multiple annotations}{Bhat et al.}

\title{Effect of latent space distribution on the segmentation of images with multiple annotations}

\author{\name Ishaan Bhat \email i.r.bhat@umcutrecht.nl \\  
	\addr Image Sciences Institute, University Medical Center Utrecht, Heidelberglaan 100, 3584 CX Utrecht, The Netherlands
	\AND
	\name Josien P.W. Pluim \email j.pluim@tue.nl \\
	\addr Department of Biomedical Engineering, Eindhoven University of Technology, Groene Loper 3, 5612 AE Eindhoven, The Netherlands
	\AND
	\name Max A. Viergever \email m.a.viergever@umcutrecht.nl \\
	\addr Image Sciences Institute, University Medical Center Utrecht, Heidelberglaan 100, 3584 CX Utrecht, The Netherlands
	\AND
	\name Hugo J. Kuijf \email h.kuijf@umcutrecht.nl \\
	\addr Image Sciences Institute, University Medical Center Utrecht, Heidelberglaan 100, 3584 CX Utrecht, The Netherlands
}

\begin{document}

\maketitle

\begin{abstract}
We propose the Generalized Probabilistic U-Net, which extends the Probabilistic U-Net by allowing more general forms of the Gaussian distribution as the latent space distribution that can better approximate the uncertainty in the reference segmentations. We study the effect the choice of latent space distribution has on capturing the variation in the reference segmentations for lung tumors and white matter hyperintensities in the brain. We show that the choice of distribution affects the sample diversity of the predictions and their overlap with respect to the reference segmentations. We have made our implementation available at \url{https://github.com/ishaanb92/GeneralizedProbabilisticUNet} 
\end{abstract}


\section{Introduction}
\label{sec:introduction}

Image segmentation may be posed as a supervised classification task with a deep learning system trained using manually created rater labels, producing a segmentation map by estimating per-voxel class probabilities. A shortcoming of standard deep learning approaches is that they produce point estimate predictions and not an output distribution from which multiple plausible predictions can be sampled~\citep{kendall_what_2017}. This disadvantage is highlighted when multiple segmentations per image are available, which standard deep learning approaches cannot leverage. Inter-rater variability reflects the disagreement among raters and ambiguity present in the image, and ideally, a supervised learning approach needs to reflect this uncertainty for unseen test cases~\citep{jungo_effect_2018}. 

The uncertainty present in a prediction can be decomposed into uncertainty originating from the model (epistemic) and uncertainty originating from the data (aleatoric)~\citep{armen_der_kiureghian_aleatory_2009}. Epistemic uncertainty can be reduced by making more data available to train a classifier, but aleatoric uncertainty is irreducible~\citep{kendall_what_2017}.
Bayesian techniques to estimate epistemic uncertainty in deep learning models usually involve producing multiple outputs by sampling model parameters from an approximating distribution~\citep{galmcd, blundell_weight_2015, kingma_variational_2015, pmlr-v80-teye18a}. The most popular among these, MC-Dropout~\citep{galmcd}, estimates pixel-wise uncertainty using dropout~\citep{srivastava_dropout_2014} to sample model parameters over multiple passes. Where multiple segmentations are available, it has been shown that regions of high estimated uncertainty reflect human disagreement~\citep{jungo_effect_2018, zhou_exploring_2019, shen_improving_2019}. While this technique is sufficient to produce good pixel-wise uncertainty estimations, it may produce spatially inconsistent segmentations~\citep{kohl_probabilistic_2018}.

Techniques that are not explicitly Bayesian in nature, like model ensembles~\citep{lakshminarayanan_simple_2017}, have been a popular choice to obtain good quality segmentations and well-calibrated epistemic uncertainty estimates~\citep{kamnitsas_ensembles_2018, mehrtash2020}. However, training a model ensemble has a large computational overhead. In \citet{rupprecht_learning_2017} the authors propose a network architecture with multiple outputs (or M-Heads) that share a common backbone network, which can mimic an ensemble with a lower compute footprint. Thus, the model ensemble and M-Heads are not scalable because the number of models (or M-Heads) are fixed during training. Furthermore, the outputs produced by ensembles may be unable to capture the diversity in the reference segmentations and fail to produce rare variants in the output distribution~\citep{kohl_probabilistic_2018}.  

Techniques that estimate aleatoric uncertainty use an approximating distribution for the input~\citep{wang_aleatoric_2019}, output~\citep{kendall_what_2017, monteiro_stochastic_2020}, or intermediate layers (latent space)~\citep{sohn_learning_2015, kohl_probabilistic_2018}. Samples drawn from this approximating distribution are used to quantify the uncertainty. 

Test-time augmentation (TTA)~\citep{wang_aleatoric_2019} is an uncertainty estimation technique that estimates aleatoric uncertainty by generating multiple plausible outputs using transformed versions of the input. In \citet{kendall_what_2017} the authors induce a Gaussian distribution over the output feature map and estimate the mean and standard deviation for each logit. Since there is no correlation between pixels, the samples produced are not realistic and the quantified uncertainty is poor~\citep{jungo_analyzing_2020}. To address this short-coming, \citet{monteiro_stochastic_2020} use a Gaussian distribution with a low-rank parameterization of the covariance matrix to model the joint distribution over the output logits, to produce diverse and plausible outputs.  

In \citet{kohl_probabilistic_2018} the authors combine the conditional variational autoencoder~\citep{sohn_learning_2015} framework with the popular U-Net~\citep{ronneberger_u-net_2015} architecture to create the Probabilistic U-Net to estimate aleatoric uncertainty by leveraging labels from multiple annotators. Different plausible variants of the prediction are computed by sampling from the (learned) latent space and combining this sample with the highest resolution feature map of the U-Net. 

In \citet{hu_supervised_2019} the authors extend the Probabilistic U-Net to capture epistemic uncertainty by using variational dropout~\citep{kingma_variational_2015} over the model weights. The PHISegNet~\citep{baumgartner_phiseg_2019} shows a further improvement in sample diversity by using a series of hierarchical latent spaces from which samples are combined with the U-Net feature maps at different resolutions.

The Probabilistic U-Net and its hierarchical variants use an axis-aligned Gaussian distribution to model the distribution over the latent space. However, techniques such as normalizing flows~\citep{rezende_variational_2016}, which convert simple distributions into complex ones via invertible transformations, have been shown to be a promising alternative to model the latent space distribution~\citep{liu_uncertainty_2020}.

Given the success of the Probabilistic U-Net (and its hierarchical variants) at estimating the aleatoric uncertainty~\citep{kohl_probabilistic_2018, baumgartner_phiseg_2019}, and the improvement shown by methods using more complex distributions to approximate the distribution over the latent space~\citep{liu_uncertainty_2020}, we propose the Generalized Probabilistic U-Net that can use more general forms of the Gaussian distribution to model the latent space distribution to better approximate the uncertainty in the reference segmentations.

In this paper, we extend our previous work~\citep{bhat_generalized_2022} with the following additions:
\begin{itemize}
    \item Inclusion of an additional dataset of white matter hyperintensities in the brain~\citep{kuijf2019}
    \item Estimation of the covariance matrix of the latent space distribution using a low-rank factor matrix (Section \ref{sec:fclr})
    \item Inclusion of popular uncertainty estimation techniques and additional experiments (Section \ref{subsec:eval})
\end{itemize}

Section \ref{sec:methods} provides background on variational inference (Section \ref{sec:cvae}) and describes our additions to the Probabilistic U-Net framework to allow more general forms of the Gaussian distribution to be used as the latent space distribution (Section \ref{sec:contribution}). Section \ref{sec:exp} describes the datasets used (Section \ref{subsec:data}), neural network training (Section \ref{subsec:nn_training}), and the evaluation metrics used to quantify performance (Section \ref{subsec:eval}). In Section \ref{sec:rnd}, we present our results and discuss their implications.

\section{Methods}
\label{sec:methods}
\subsection{Background}
\label{sec:cvae}
Variational inference (VI)~\citep{wainwright_graphical_2007} is a tool to approximate complex probability distributions in a scalable and computationally efficient manner. Variational autoencoders (VAEs)~\citep{Kingma2014, pmlr-v32-rezende14}, a class of deep latent variable models, were developed to perform generative modelling with possible one-to-many mappings. They do so by learning a distribution over the latent variables using variational inference, and sample from this distribution to generate multiple plausible outputs. In \citet{sohn_learning_2015} the authors created a conditional variant (cVAE) to handle structured prediction problems like image segmentation.

\begin{figure*}[htb]
\centering
\begin{subfigure}[t]{0.45\linewidth}
\includegraphics[width=1.0\linewidth]{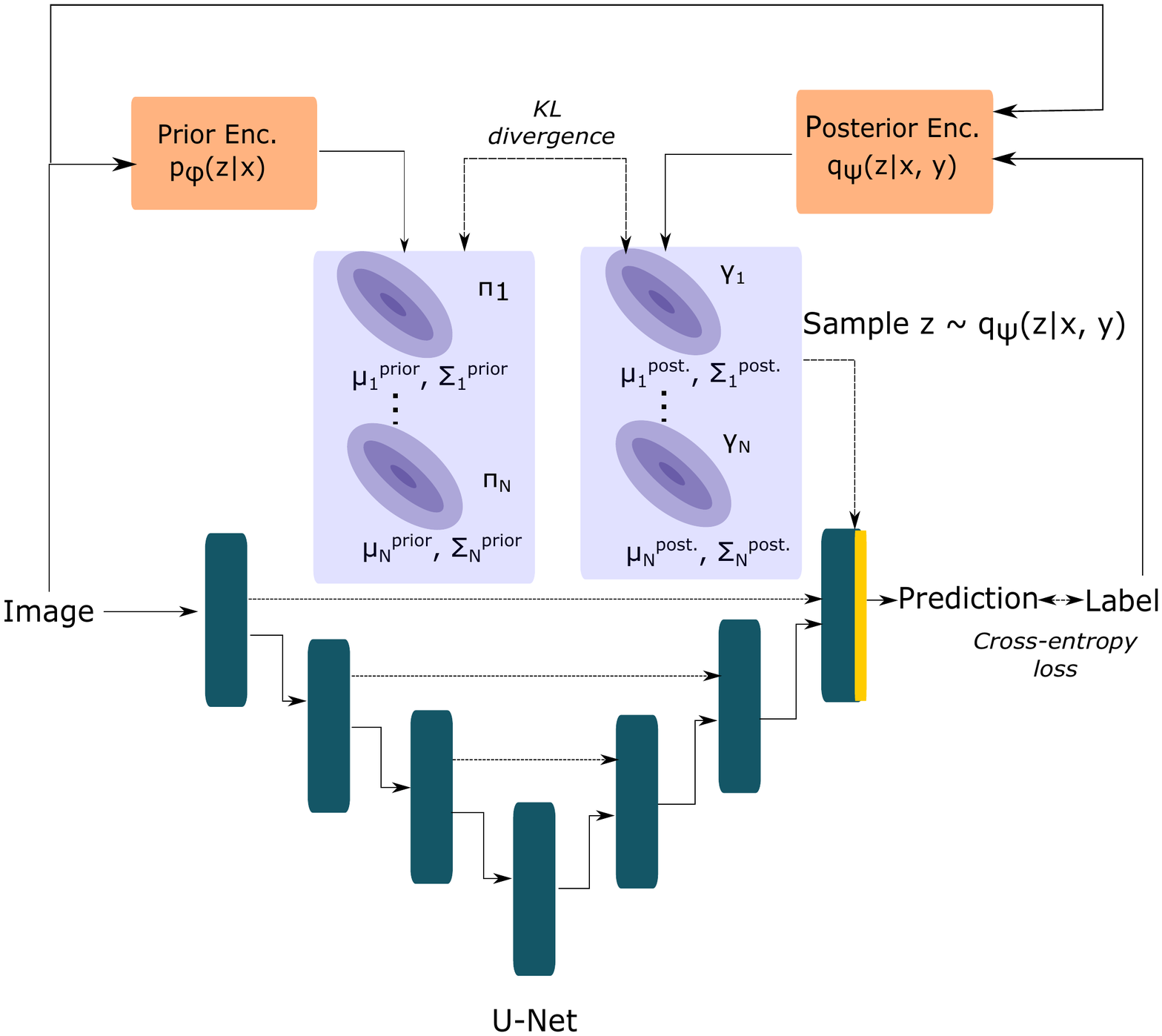}
  \caption{Training}
  \label{fig:punet_train}
\end{subfigure}
 \hspace{1em}
\begin{subfigure}[t]{0.45\linewidth}
  \includegraphics[width=1.0\linewidth]{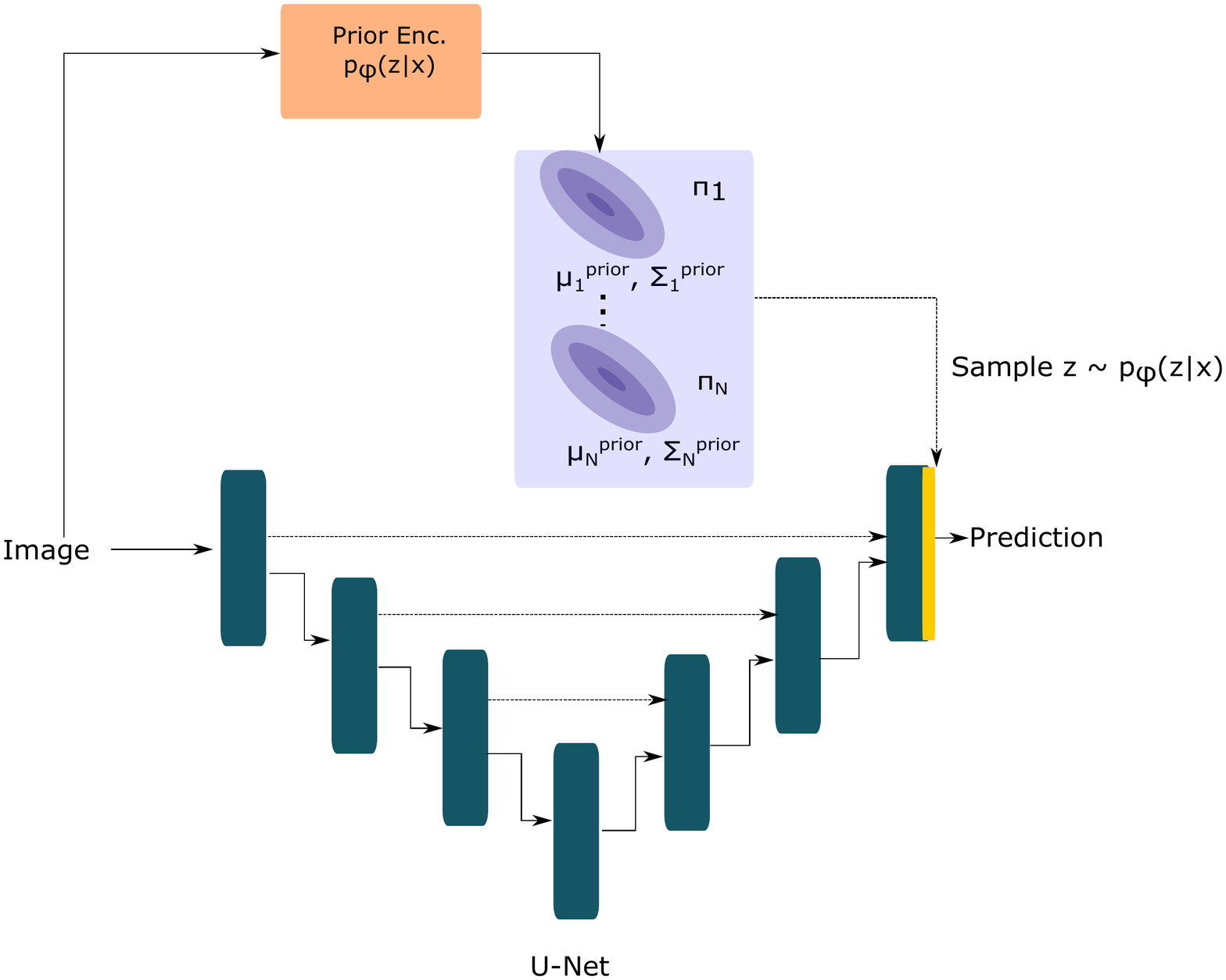}
  \caption{Inference}
  \label{fig:punet_test}
\end{subfigure}
\caption{Generalized Probabilistic U-Net framework. In Figure \ref{fig:punet_train}, the model learns prior and posterior distribution parameters $\{\mu_i^{\mathrm{prior}}, \Sigma_i^{\mathrm{prior}}, \pi_{i}\}_{i=1}^N$ and $\{\mu_i^{\mathrm{post.}}, \Sigma_i^{\mathrm{post.}}, \gamma_{i}\}_{i=1}^N$. In the most general case, the prior and posterior distributions are modelled as a mixture of $N$ Gaussians, which can be reduced to a single Gaussian for $N=1$. Similarly, different variants of the multivariate Gaussian distribution can be modelled by constructing the covariance matrix in the different ways described in Section \ref{sec:contribution}. During inference, the posterior encoder is discarded, and different plausible outputs can be computed by sampling from the prior distribution and combining this sample with the last U-Net layer.}
\label{fig:punet}
\end{figure*}

In a medical image segmentation problem, we typically have an image $x \in \mathbb{R}^{H \times W}$ and label $y \in \{0, 1\}^{H \times W}$ and the goal is to produce accurate and coherent samples from the distribution $p_{\theta}(y | x)$ using a deep neural network parameterized by $\theta$. To enable tractable sampling from $p_{\theta}(y|x)$, simple  distributions $p_{\phi}(z|x)$ and $q_{\psi}(z|x, y)$ are used as approximations for the true prior and posterior distributions over the latent space respectively. The distributional parameters for $q_{\psi}(z|x, y)$ and $p_{\phi}(z|x)$ are learnt by neural (sub-)networks parameterized by $\psi$ and $\phi$ respectively.

The training objective for the cVAE is minimizing the KL-divergence between the true and approximating posterior distributions~\citep{sohn_learning_2015}:

\begin{equation}
    \label{eq:elbo}
\mathbb{L}_{Tr}(\theta, \phi, \psi; x, y) = -\mathbb{E}_{q_{\psi}(z|x, y)}[\mathrm{log}\,p_{\theta}(y|x, z)] + \beta*\mathrm{KL}(q_{\psi}(z|x, y) || p_{\phi}(z|x))
\end{equation}

In Equation \ref{eq:elbo}, the first term is the expected value of the standard cross-entropy loss computed under the posterior distribution. The second term is the Kullback-Leibler (KL) divergence between the prior distribution, $p_{\phi}(z|x)$, and the posterior distribution, $q_{\psi}(z|x, y)$. This term acts as a regularization term that pulls the prior and posterior distribution closer to each other by penalizing their differences~\citep{kohl_probabilistic_2018}. This is desirable since the posterior network is discarded during test-time. The $\beta$ term controls the trade-off between the segmentation and KL-loss terms~\citep{Higgins2017betaVAELB}.

The reparameterization trick~\citep{Kingma2014} is used to generate samples from $q_{\psi}(z|x, y)$ in a computationally efficient manner. This trick is used to obtain an unbiased estimate of the gradient of the expectation term $\mathbb{E}_{q_{\psi}(z|x, y)}[\mathrm{log}\,p_{\theta}(y|x, z)]$ and make it differentiable with respect to parameters $\psi$. Therefore, it is convenient if the distribution chosen to approximate the latent space distribution supports the reparameterization trick.

\subsection{Contributions}
\label{sec:contribution}
The cVAE~\citep{sohn_learning_2015} uses an axis-aligned Gaussian to model the prior and posterior distributions over the latent space. The Probabilistic U-Net~\citep{kohl_probabilistic_2018} combines the VI framework of the cVAE with the U-Net~\citep{ronneberger_u-net_2015} architecture. Additionally, it makes output computation efficient by merging the sample from the latent space distribution at the last layer of the U-Net.    

We propose three extensions to this framework by using more general forms of the Gaussian distribution as choices for the latent space distributions in order to better capture the uncertainty in the reference segementations. Our generalized Probabilistic U-Net framework is shown in Figure \ref{fig:punet}.

\subsubsection{Full-covariance Gaussian}
For a matrix to be a valid covariance matrix, it must be  positive semi-definite. Since this constraint is difficult to impose while training a neural network, the covariance matrix $\Sigma$ is built using its Cholesky decomposition $L$~\citep{williams1996}:
\begin{equation*}
    \Sigma = LL^{T}
\end{equation*}
The matrix $L$ is a lower-triangular matrix with a positive-valued diagonal. By masking the upper-triangular section of the matrix and using the exponential operator to ensure positive values on the diagonal, $L$ can be directly computed by the neural network. Samples can be drawn from the full-covariance Gaussian using the reparameterization trick~\citep{kingma_introduction_2019}:
\begin{align*}
    z = \mu + L*\epsilon,\ \epsilon \sim \mathcal{N}(0, I)
\end{align*}

\subsubsection{Full-covariance Gaussian with a low-rank parameterization}
\label{sec:fclr}
A limiting factor in modelling a Gaussian with a full covariance matrix using neural networks is often the squared dependence between the size of the covariance matrix and dimensionality of the random variable.
In \citet{barber_ensemble_1998} the authors use a low-rank factor matrix $P$ and a diagonal matrix $D$ to build the covariance matrix used to model the posterior distribution over the neural network weights. We use this technique to build the covariance matrix $\Sigma$ for our latent space distribution:
\begin{equation*}
    \Sigma = PP^{T} + D
\end{equation*}

For a $N$-dimensional Gaussian distribution, $D$ is a $N$-dimensional vector with the diagonal elements and $P$ is a $N \times R$ factor matrix, where $R$ is a hyperparameter that defines the rank of the parameterization. The $P$ and $D$ matrices are computed as outputs of the posterior and prior encoders. This low-rank parameterization also supports the reparameterization trick for sampling via the Cholesky decomposition of $\Sigma$~\citep{monteiro_stochastic_2020}.

\subsubsection{Modelling of mixture of Gaussians}
Any arbitrary (continuous) distribution can be modelled by a mixture of a sufficient number of Gaussians, with appropriate mixture weights~\citep{bishop2006}. Mixtures of Gaussians have been used to model posterior distributions in variational autoencoders for clustering tasks~\citep{jiang_variational_2017, kopf_mixture--experts_2021}.

For example, the posterior distribution\footnote{This subsection holds true for the cVAE prior distribution as well. The only difference is the dependence on the label, $y$, is dropped.} $q_{\psi}(z | x, y)$ can be modelled as a mixture of $N$ Gaussians as follows:
\begin{equation*}
    q_{\psi}(z | x, y) = \sum_{i=1}^{N} \gamma_i \mathcal{N}(\mu_i(x, y), \Sigma_i(x, y))
\end{equation*}
The individual Gaussians in the mixture, $\mathcal{N}(\mu_i(x, y), \Sigma_i(x, y))$, are called the component distributions and the weights for the component distributions, $\{\gamma_i\}_{i=1}^N$, are the mixing coefficients. The individual Gaussians can be modelled using a diagonal or full covariance matrix. For the distribution to be valid, the mixing coefficients must be greater than or equal to $0$ and sum to $1$. Therefore, a categorical distribution can be used to define the mixture distribution.

As explained in Section \ref{sec:cvae}, a distribution chosen to model the posterior must support differentiable sampling. To sample from a mixture of Gaussians, a component index is sampled from the categorical distribution defined by the mixture coefficients and then a value is sampled from the corresponding component (Gaussian) distribution. 
\begin{gather*}
    i \sim \mathrm{Cat}(N;\gamma) \\
    z \sim \mathcal{N}(\mu_i(x, y), \Sigma_i(x, y))
\end{gather*}
The second step in the sampling process is differentiable and supports the reparameterization trick, however, the first step, that is, sampling from a categorical distribution is not differentiable. To make sampling fully differentiable, we used the Gumbel-Softmax (GS) distribution\citep{jang_categorical_2017, maddison_concrete_2017} to model the mixture distribution. The Gumbel-Softmax distribution is a continuous relaxation (defined by the temperature parameter, $\tau$) of the discrete categorical distribution, that supports differentiable sampling via the reparameterization trick. To obtain a discrete component index, we perform Straight-Through(ST) sampling~\citep{jang_categorical_2017}, where we used the $\mathrm{argmax}$ operation in the forward pass and used the continuous relaxation in the backward pass while computing gradients used to update model parameters during training. With this, we used the following two-step process to sample from the mixture of Gaussians:
\begin{gather*}
    i \sim \mathrm{GS}(N;\gamma, \tau) \\
    z \sim \mathcal{N}(\mu_i(x, y), \Sigma_i(x, y))
\end{gather*}
Unlike for a pair of Gaussians, the KL divergence for a pair of Gaussian mixtures does not have a closed-form expression. We estimated the KL divergence in Equation \ref{eq:elbo} via Monte Carlo sampling, which provides a good approximation~\citep{hershey2007}.

\begin{figure*}[htb]
\centering
\begin{subfigure}[t]{0.45\linewidth}
\includegraphics[width=1.0\linewidth]{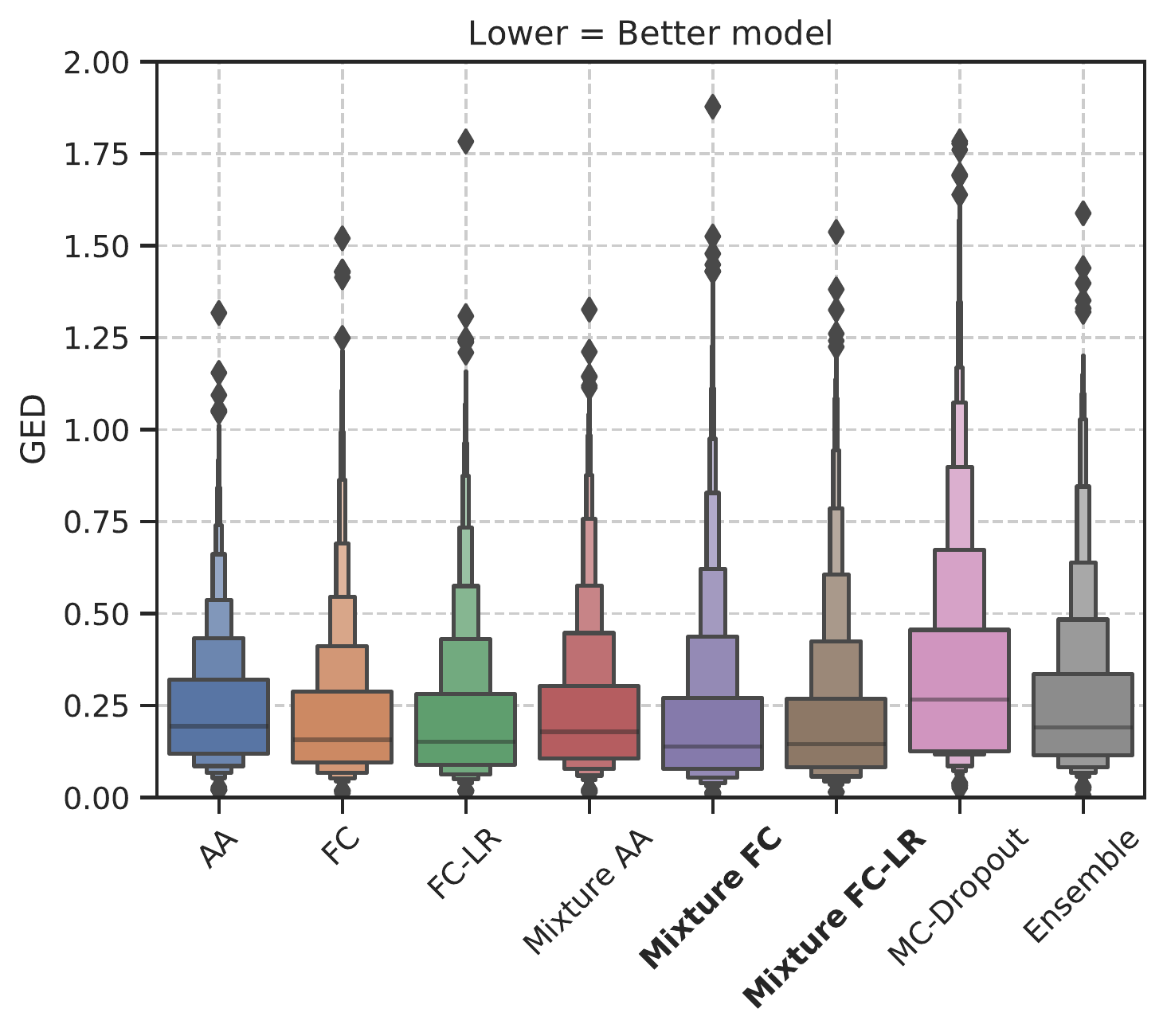}
  \caption{LIDC-IDRI}
  \label{fig:ged_lidc}
\end{subfigure}
 \hspace{1em}
\begin{subfigure}[t]{0.45\linewidth}
  \includegraphics[width=1.0\linewidth]{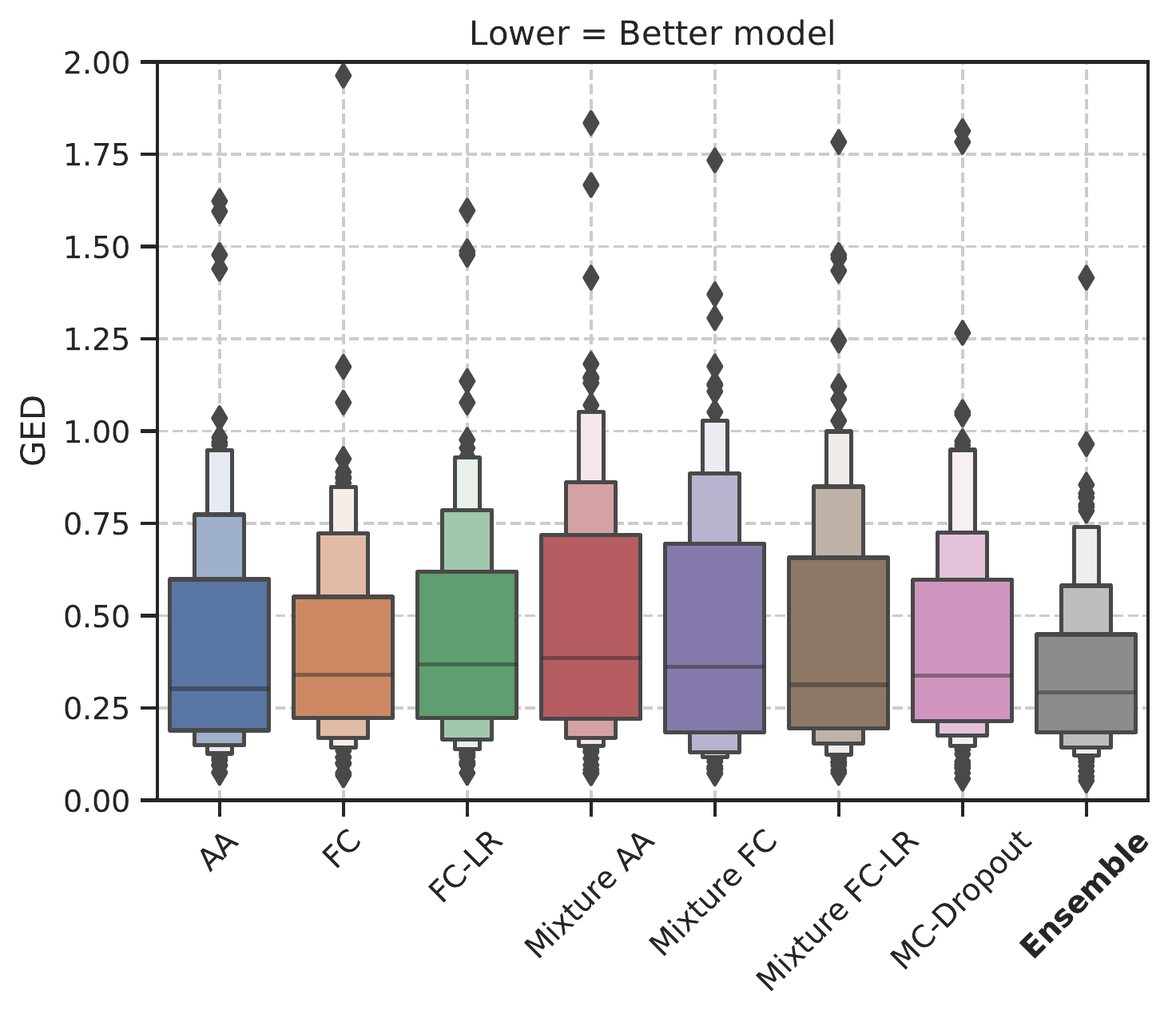}
  \caption{WMH}
  \label{fig:ged_wmh}
\end{subfigure}
\caption{Trends in the GED metric for LIDC-IDRI and WMH datasets using letter-value plots. Best performing configurations in bold (multiple bold configurations indicate lack of a statistically significant difference). The median is shown by a horizontal line segment, and the innermost box is drawn at the upper and lower fourths, that is, 25th and 75th percentiles. The other boxes are drawn at the upper and lower eighths, sixteenths and so on. The GED was computed using $16$ prediction samples.}
\label{fig:ged_trends}
\end{figure*}

\begin{table*}[htp]
\centering
\caption{Neural network hyperparameters for LIDC-IDRI and WMH datasets. The number of U-Net encoder and decoder blocks were fixed at $3$, and the filter depths used were $32, 64, 128$ for the first, second and third blocks respectively. The bottleneck layer had a filter depth of $512$. The numbers below each of the hyperparameters indicate the search space, with square brackets indicating a full grid search among the values between parenthesis, and curved brackets indicating random sampling in the range of values between parenthesis.}
\resizebox{1.0\textwidth}{!}{\begin{tabular}{@ {} l@{\hspace{1cm}} l@{\hspace{1cm}} l@{\hspace{1cm}} l@{\hspace{1cm}} l @ {\hspace{1cm}} l @ {\hspace{1cm}} l @ {\hspace{1cm}}}
\toprule
Model       & \multicolumn{6}{c}{Hyper-parameters (LIDC-IDRI / WMH)} \\ \midrule
              & Latent space dimension    & $\beta$ & Rank & Mixture components & Temperature & Dropout rate  \\
              & $[2, 4, 6, 8]$ & $[1, 10, 100]$ & $[1, 7]$ & $(1, 10)$ & $(0.1, 0.5)$ & $[0.1, 0.3, 0.5]$ \\
              \cmidrule{2-7}
AA & 6 / 4 & 1 / 1 & - / - & - / - & - / - & - / - \\
FC   & 2 / 2 & 1 / 1 & - / - & - / - & - / - & - / - \\
FC-LR & 2 / 2 & 1 / 1 & 1 / 1 & -/- & - / - & - / - \\
AA Mixture  & 4 / 4 & 1 / 1 & - & 5 / 3 & 0.36 / 0.20 & - / - \\
FC Mixture      & 2 / 4 & 1 / 1 & - & 9 / 4 & 0.28 / 0.22 & - / - \\
FC-LR Mixture   & 4 / 8 & 1 / 1 & 2 / 7 & 2 / 9 & 0.20 / 0.17 & - / -  \\
MC-Dropout & - / - & - / - & - / - & - / - & - / - & 0.3 / 0.3  \\
\bottomrule
\end{tabular}}

\label{tab:hyper}
\end{table*}

\section{Experiments}
\label{sec:exp}
\subsection{Data}
\label{subsec:data}
\subsubsection{LIDC-IDRI}

The LIDC-IDRI dataset~\citep{lidc2011} consists of $1018$ thoracic CT scans with lesion segmentations from four raters. Like \citet{kohl_probabilistic_2018} and \citet{baumgartner_phiseg_2019}, we use a pre-processed version of the dataset, consisting of $128 \times 128$ patches containing lesions.

We obtained a total of $15,096$ patches. We used a $60:20:20$ split to randomly partition the dataset into $9058,\,3019,$ and $3019$ patches for training, validation, and test respectively. The intensity of the patches was scaled to $[0, 1]$ range and no data augmentation was used. We used the publicly available version of the pre-processed dataset from \url{https://github.com/stefanknegt/Probabilistic-Unet-Pytorch}.

\subsubsection{White matter hyperintensities (WMH)}

The WMH dataset~\citep{kuijf2019} consists of $60$ MR images collected from three centers: University Medical Center (UMC) Utrecht ($20$ patients), VU University Medical Center (VU) Amsterdam ($20$ patients), both in the Netherlands, and the National University Health System (NUHS) ($20$ patients) in Singapore. 

For each patient, a 3D T1-weighted and a 2D multi-slice FLAIR image were provided. For each patient, the 3D T1-weighted image was registered to the FLAIR image using the elastix~\citep{elastix2010} toolbox to create a two-channel image. A brain mask was created from the FLAIR image using the ROBEX~\citep{iglesias2011robust} technique, which was applied to the T1-weighted image as well.  All $60$ images had white matter hyperintensities annotated by three raters. 

We used a $80:10:10$ split to get $48,\,6$, and $6$ patients for training, validation, and test respectively. From each patient MR image, we only used slices that have at least one non-empty segmentation. All images and labels were padded to be $256\times256$ in dimension. Each image was normalized using the (channel-wise) mean and standard deviation.  
During training, we used random flips, random rotation, contrast, and brightness manipulation to perform data augmentation. 

\subsection{Neural network training}
\label{subsec:nn_training}
For all datasets and models, we used the ADAM~\citep{kingma_adam:_2015} optimizer with a learning rate of $10^{-4}$. Training was stopped when the validation loss did not improve for more than 20 epochs, and the model parameters with the minimum validation loss were saved. During each iteration, random sampling was performed to select a rater and the corresponding ground truth segmentation was chosen to calculate the training loss and perform parameter updates. We chose hyperparameters for both datasets and the different configurations of the Generalized Probabilistic U-Net via a random grid search strategy. The hyperparameters for all datasets and models are shown in Table \ref{tab:hyper}.

To ensure comparability between models with different latent space distributions, we maintain the same number of layers for the U-Net, the prior encoder, and the posterior encoder. Furthermore, in accordance with the original Probabilistic U-Net architecture, the prior and posterior encoder have the same architecture as the U-Net encoder. Each convolution block in the model consisted of convolution with a $3 \times 3$ kernel, a ReLU nonlinearity and batch normalization~\citep{ioffe15}. Downsampling is performed via average pooling and upsampling  is performed via bilinear interpolation.

The code for the neural network training and inference was developed using the PyTorch~\citep{pytorch2019} library.

\subsection{Evaluation}
\label{subsec:eval}
\begin{figure*}[htb]
\centering
\begin{subfigure}[t]{0.49\linewidth}
\includegraphics[scale=0.3]{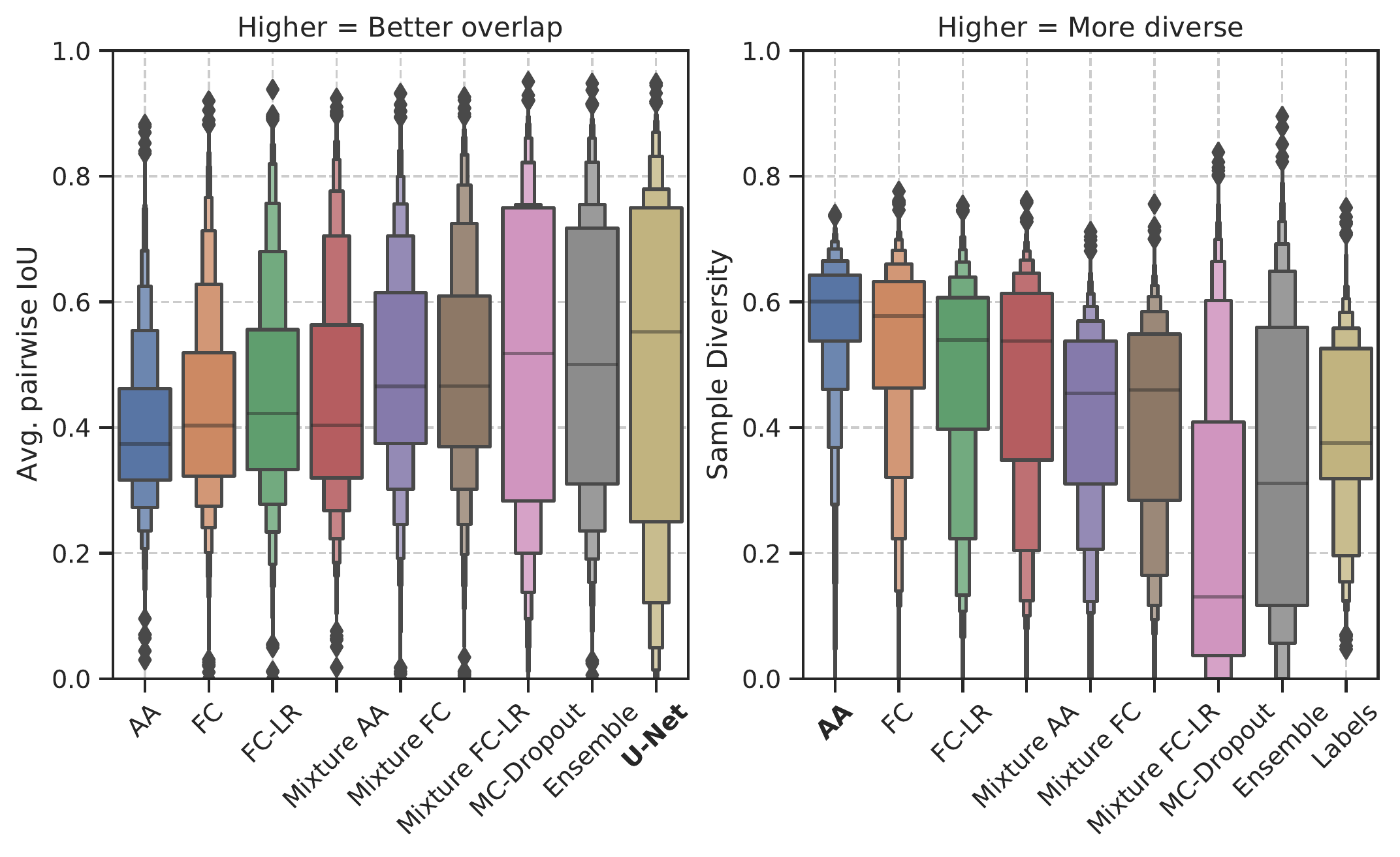}
  \caption{LIDC-IDRI}
  \label{fig:ged_breakup_lidc}
\end{subfigure}
\begin{subfigure}[t]{0.49\linewidth}
  \includegraphics[scale=0.3]{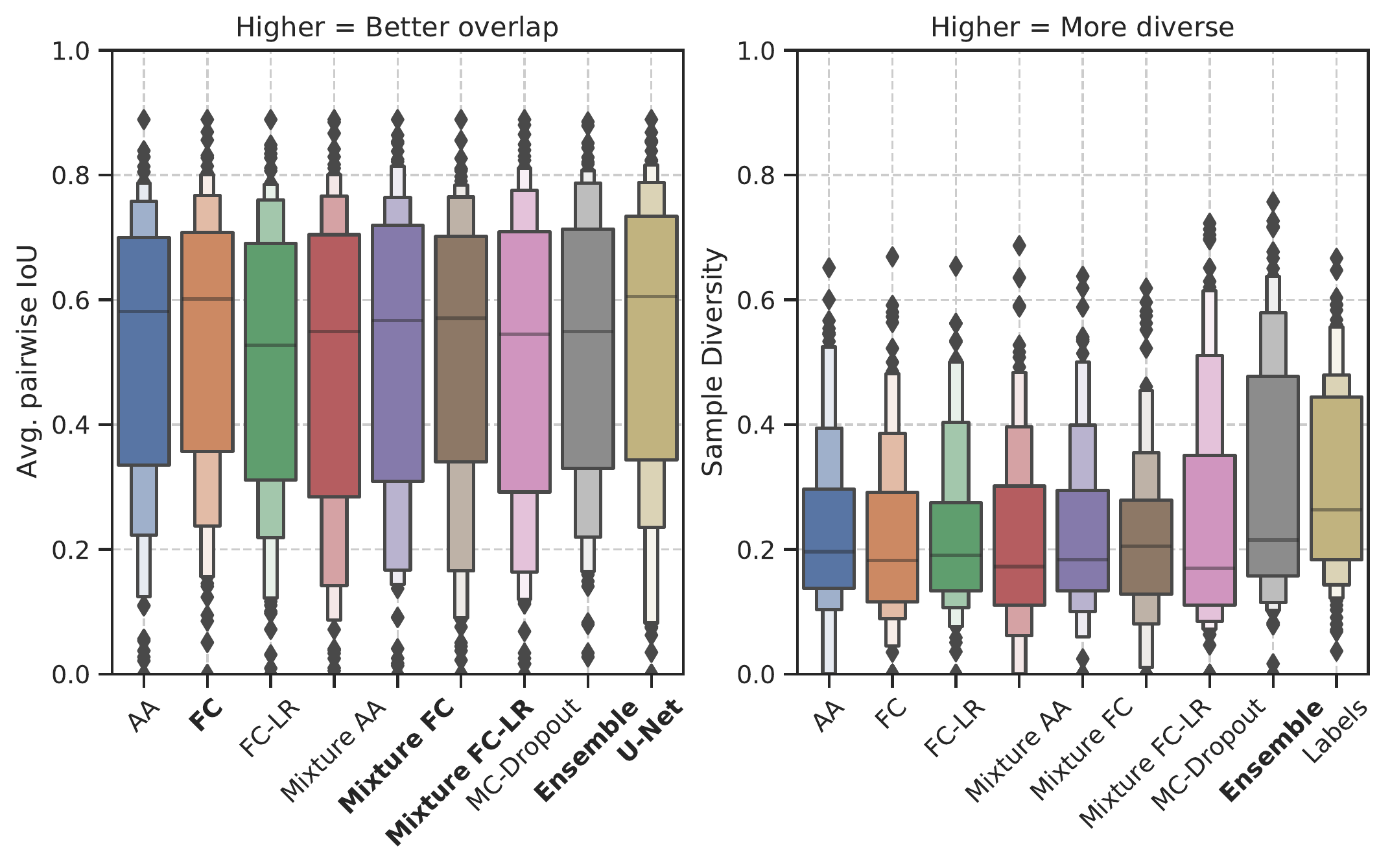}
  \caption{WMH}
  \label{fig:ged_breakup_wmh}
\end{subfigure}
\caption{GED divided into average IoU and sample diversity terms for the LIDC-IDRI and WMH datasets. Best performing configurations in bold (multiple bold configurations indicate lack of a statistically significant difference).}
\label{fig:ged_breakup}
\end{figure*}

We study different configurations of the Generalized Probabilistic U-Net by using different distributions (Section \ref{sec:contribution}) to model the latent space distributions. Additionally, we included the popular uncertainty estimation techniques MC-Dropout~\citep{galmcd} and model ensembles~\citep{lakshminarayanan_simple_2017}. We included the following models in our experiments:
\begin{itemize}
    \item Axis-aligned Probabilistic U-Net (AA) \citep{kohl_probabilistic_2018}
    \item Full covariance Probabilistic U-Net (FC)
    \item Full covariance Probabilistic U-Net [Low-rank parameterization] (FC-LR)
    \item Mixture AA Probabilistic U-Net
    \item Mixture FC Probabilistic U-Net
    \item Mixture FC-LR Probabilistic U-Net
    \item MC-Dropout~\citep{galmcd}
    \item Model ensemble~\citep{lakshminarayanan_simple_2017}
\end{itemize}

\begin{figure*}[!htbp]
\centering
\begin{subfigure}[t]{0.8\linewidth}
\includegraphics[width=1.0\linewidth,height=0.6\linewidth]{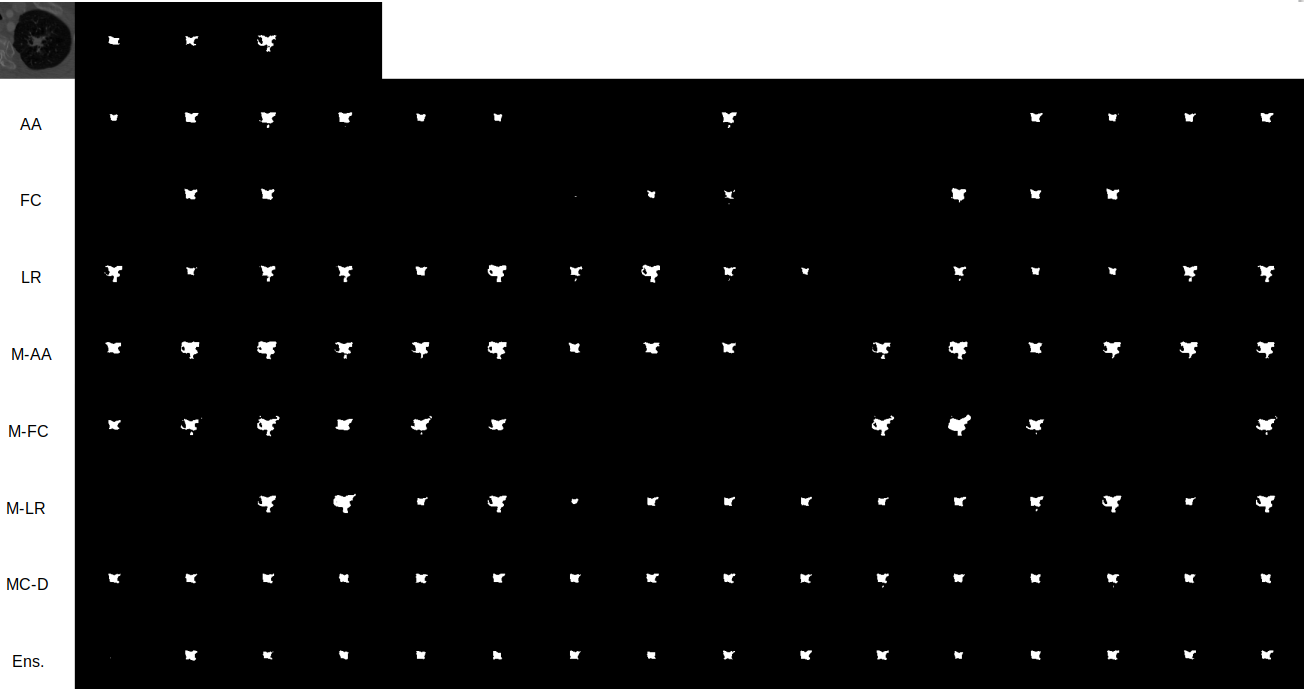}
  \caption{LIDC-IDRI}
  \label{fig:lidc_tile}
\end{subfigure}
\begin{subfigure}[t]{0.8\linewidth}
  \includegraphics[width=1.0\linewidth,height=0.7\linewidth]{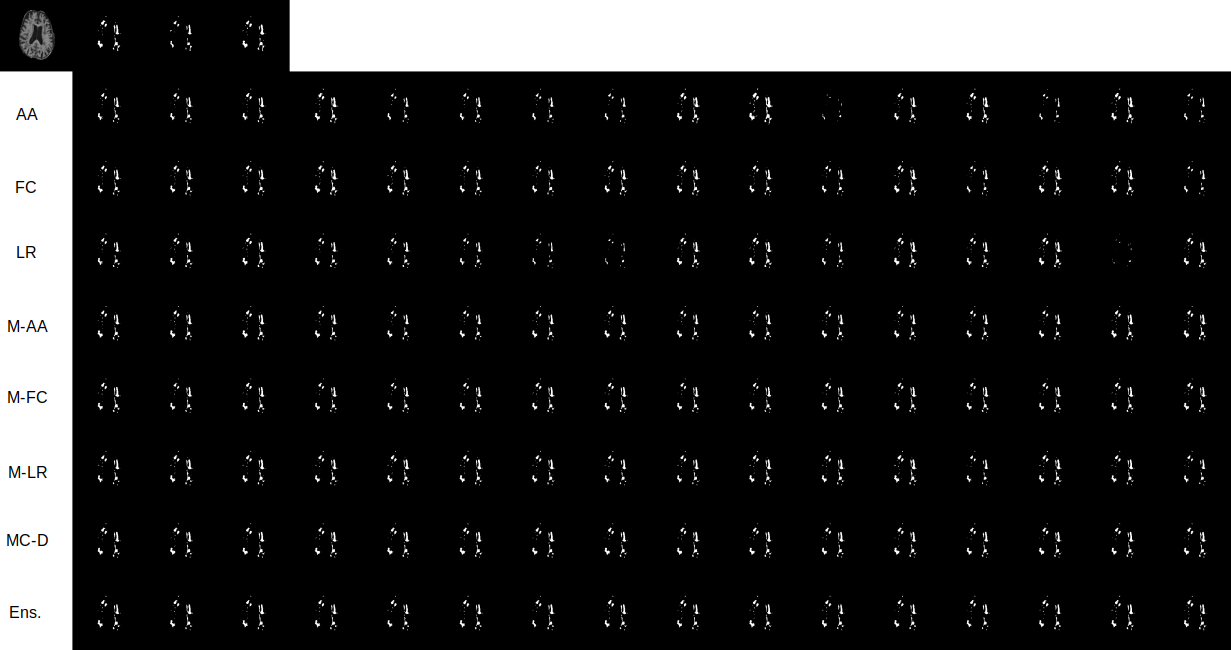}
  \caption{WMH}
  \label{fig:wmh_tile}
\end{subfigure}
\caption{Example image, labels, and predictions for both datasets. The first row contains the image and the ground truth labels made by the different annotators. The following rows show $16$ samples drawn from the prediction distribution (used to compute the GED).}
\label{fig:image_tiles}
\end{figure*}

\begin{figure*}[htb]
\centering
\includegraphics[width=1.0\linewidth]{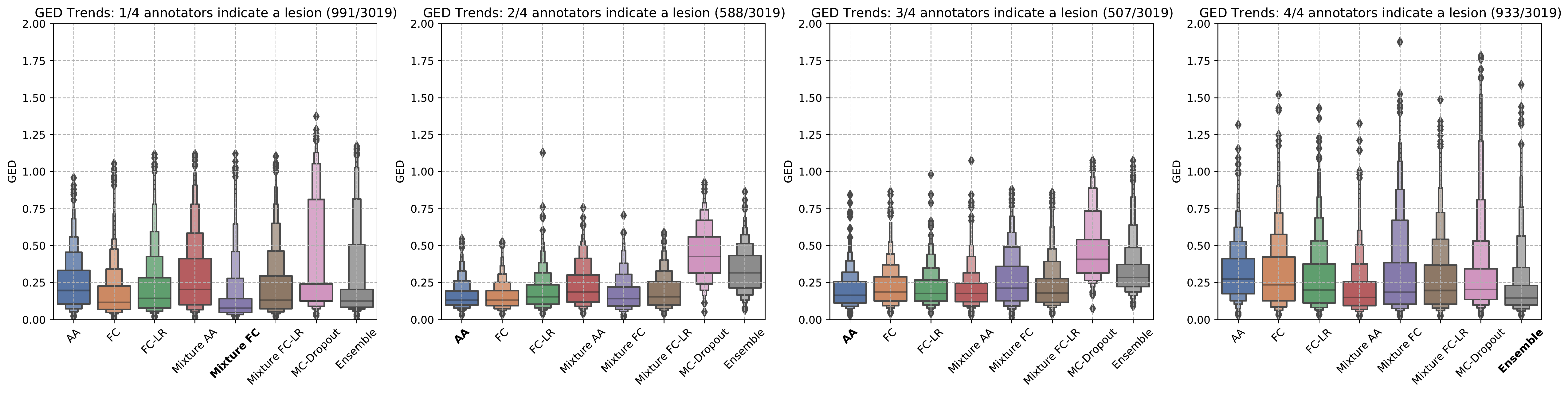}
\caption{GED trends within the LIDC-IDRI test-set grouped by rater agreement. Best performing configurations in bold.}
\label{fig:ged_bucket}
\end{figure*}

To ensure a fair comparison between the Generalized Probabilistic U-Net configurations and the rest, MC-Dropout and model ensemble used the same U-Net architecture as the Probabilistic U-Net variants. The model ensemble was created by training $16$ models independently using different random seeds. The dropout rate for the MC-Dropout model was selected based on minimum validation loss. The dropout rate chosen for the WMH and LIDC-IDRI datasets was $0.3$.  

We used the \emph{generalized energy distance} (GED) metric~\citep{kohl_probabilistic_2018} to compare how well the distribution of neural network predictions $P_s$ matched the distribution of ground truth labels $P_{gt}$.

\begin{equation}
\label{eq:ged}
D^2_{GED}(P_{gt}, P_s) =  2\mathbb{E}[d(S, Y)] - \mathbb{E}[d(S, S')] \\ - \mathbb{E}[d(Y, Y')]
\end{equation}

The distance metric between a pair of segmentations, $d(x, y)$, is defined as $1 - \mathrm{IoU}(x, y)$~\citep{kohl_probabilistic_2018, baumgartner_phiseg_2019}. In Equation \ref{eq:ged}, $S, S'$ are samples from the prediction distribution and $Y, Y'$ are samples from the ground truth distribution. The expectation terms in Equation \ref{eq:ged} are computed via Monte Carlo estimation. We used $16$ samples from the prediction distribution to compute the GED. A lower GED implies a better match between prediction and ground truth distributions.

We also studied the trends in the component terms of the GED metric. The first term $\mathbb{E}[d(S, Y)]$ signifies the extent of overlap between samples from the prediction and ground truth distribution, while the second term $\mathbb{E}[d(S, S')]$ is the sample diversity in the predictions.  As a baseline for the overlap trends, we included results obtained on the test-set using a single U-Net~\citep{ronneberger_u-net_2015}. To provide context for the sample diversity in the predictions, we included the diversity in the labels in our results. By investigating the correlation of these terms with the GED metric, we studied the interplay between overlap and diversity that lead to a better (or worse) match between distributions.

To study the effect the choice of latent space distribution has on capturing uncertainty, we partitioned the test-set of the LIDC-IDRI dataset into four subsets, based on rater agreement. The slices were assigned to one of the four subsets based on the number of raters (1-4) that annotated a lesion in it.

We studied the influence of the \emph{rank} (number of columns in $P$) hyperparameter  used to construct the covariance matrix for the FC-LR configuration of the Generalized Probabilistic U-Net. For each of the latent space dimensions in the hyperparameter search space (Table \ref{tab:hyper}), we analysed the trends in the GED for the allowed values of the rank.

\begin{figure*}[htb]
\centering
\begin{subfigure}[t]{0.8\linewidth}
\includegraphics[width=1.0\linewidth]{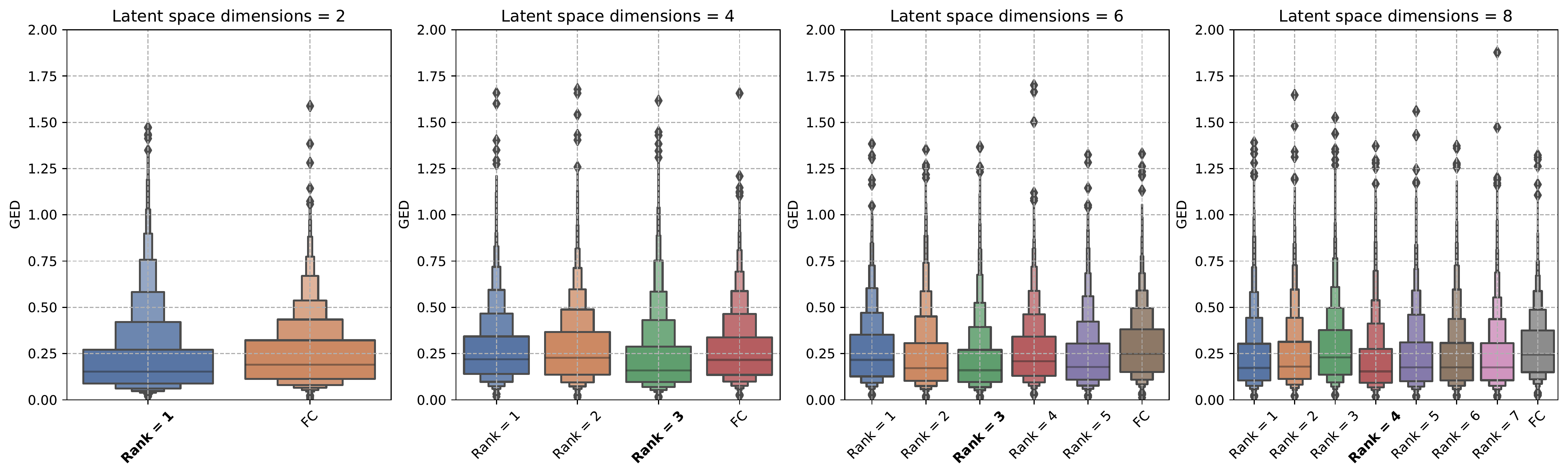}
  \caption{LIDC-IDRI}
  \label{fig:rank_lidc}
\end{subfigure}
\begin{subfigure}[t]{0.8\linewidth}
  \includegraphics[width=1.0\linewidth]{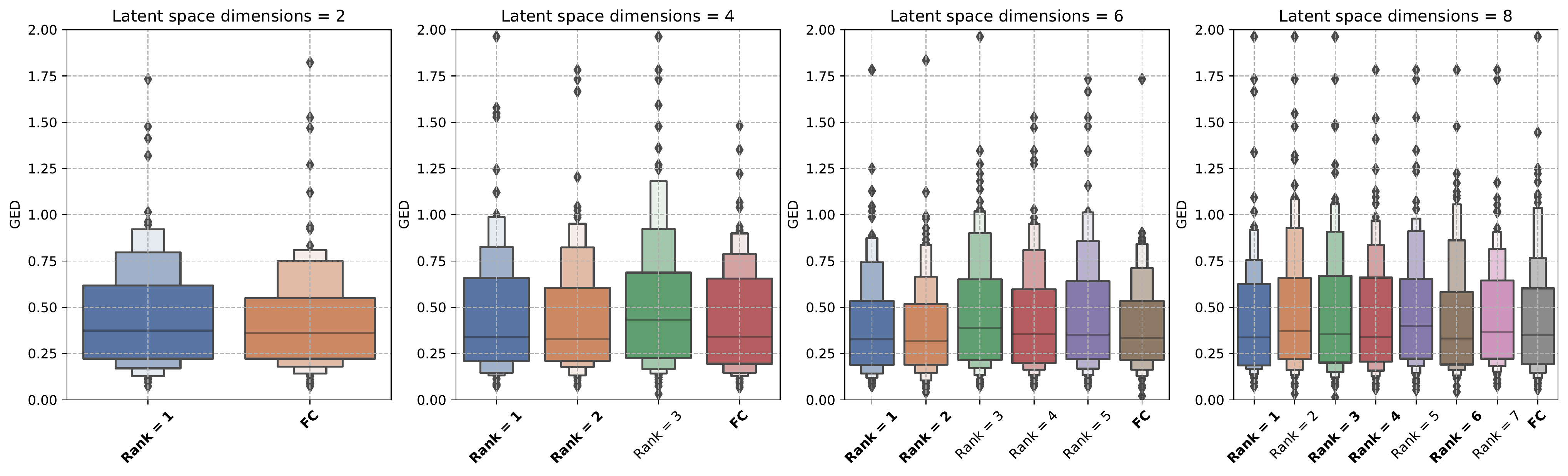}
  \caption{WMH}
  \label{fig:rank_wmh}
\end{subfigure}
\caption{Influence of the \emph{rank} hyperparameter on GED trends. Best performing configurations in bold (multiple bold configurations indicate lack of a statistically significant difference). For each latent space dimension $z$, $\mathrm{\emph{rank}} \in [1, z-1]$. We include GED trends for the full covariance (FC) configuration for each latent space dimension for comparison.}
\label{fig:rank_exp}
\end{figure*}

Lastly, we studied the effect the number of raters has on the ability of the model to predictions that accurately match the ground truth distribution. To do so, we trained each model on a sub-set of raters and evaluated their performance using segmentations from all the raters.

To check for statistical significance of differences in the metrics for models, we performed the Wilcoxon signed-rank test at a significance of $0.05$.

\begin{figure*}[htb]
\centering
\begin{subfigure}[t]{0.8\linewidth}
\includegraphics[width=1.0\linewidth]{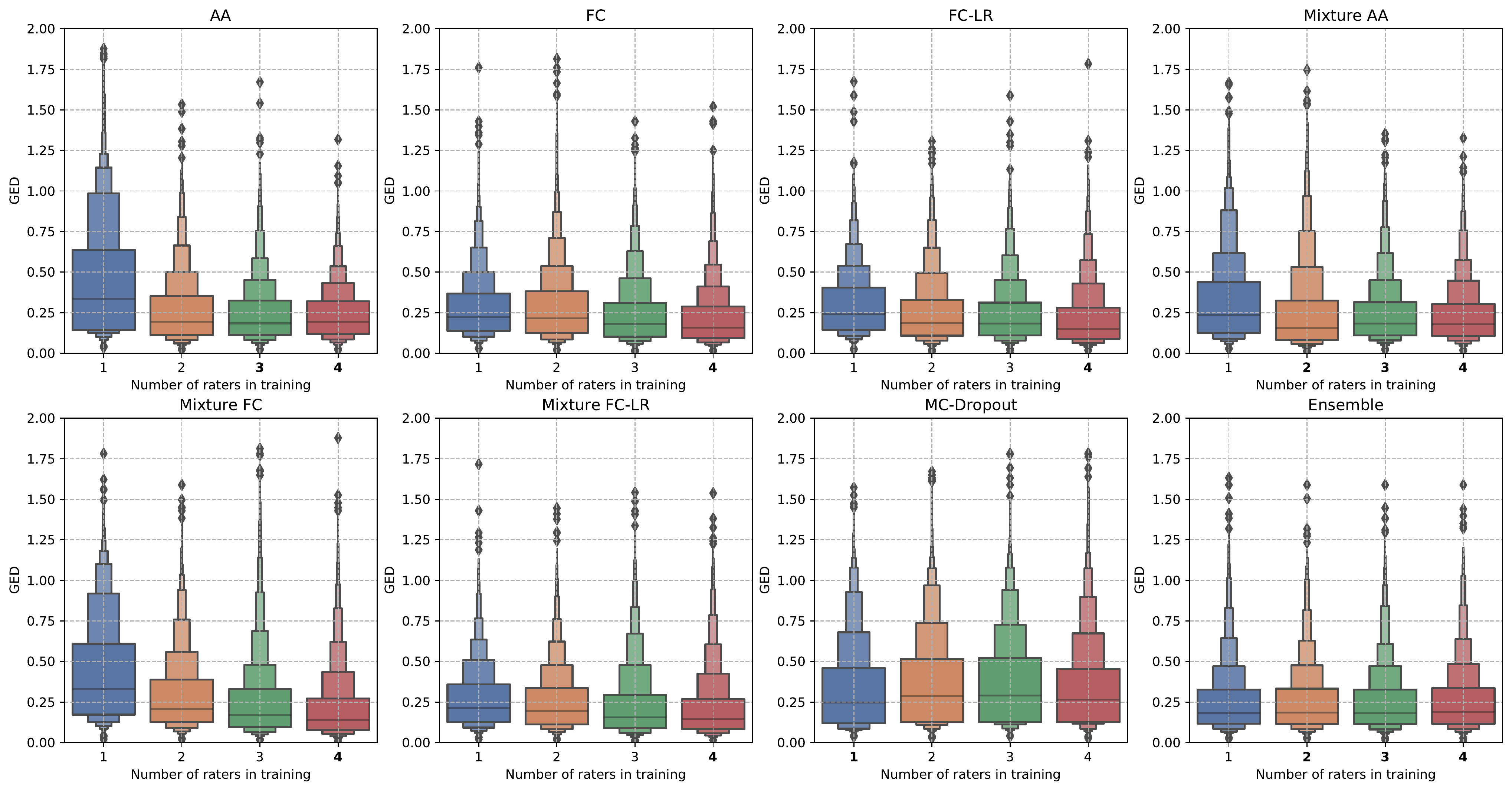}
  \caption{LIDC-IDRI}
  \label{fig:ged_raters_lidc}
\end{subfigure}
\begin{subfigure}[t]{0.8\linewidth}
  \includegraphics[width=1.0\linewidth]{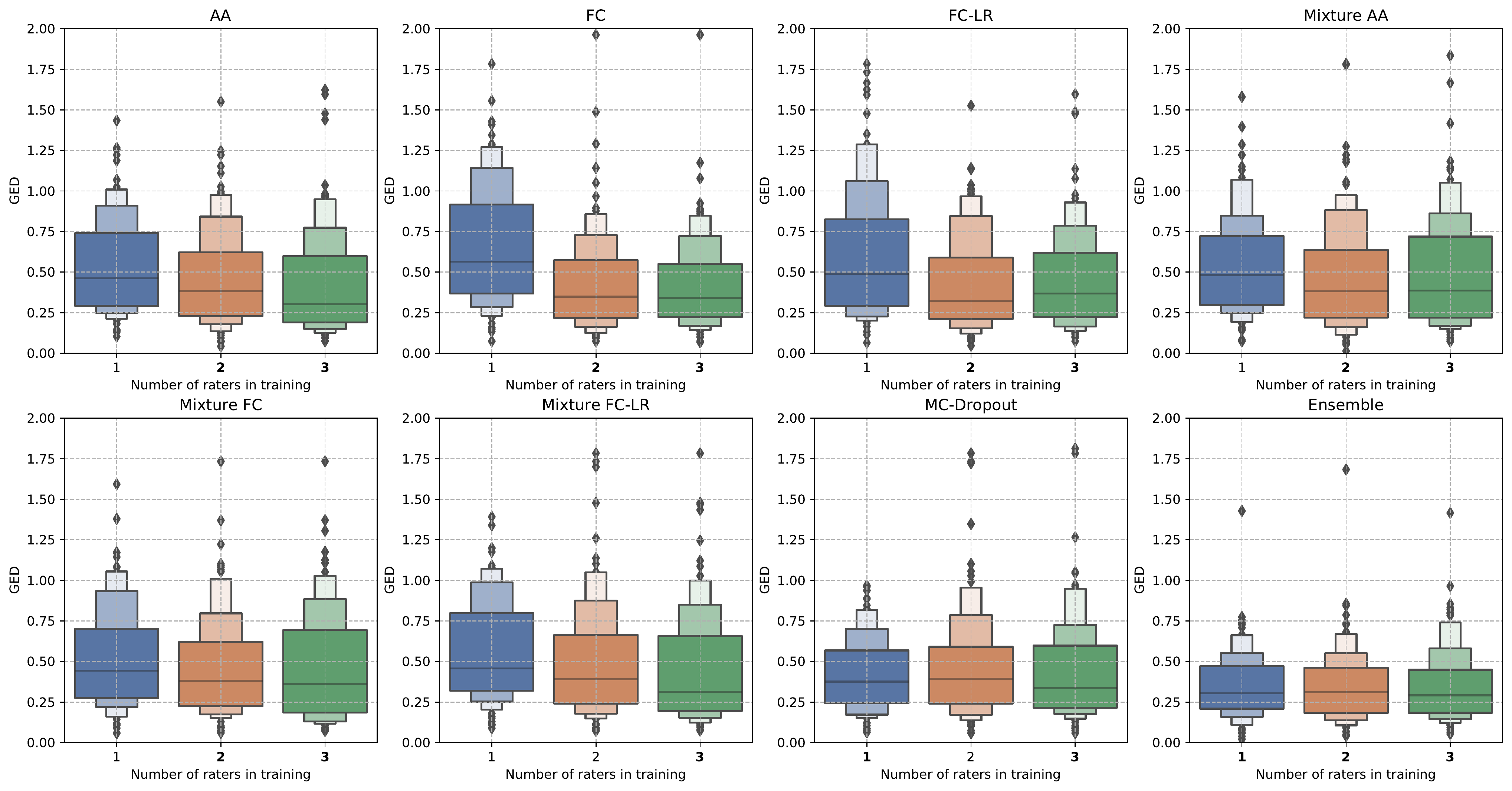}
  \caption{WMH}
  \label{fig:ged_raters_wmh}
\end{subfigure}
\caption{GED metric trends as number of raters is increased during training. During evaluation, segmentations by all raters were used. Best performing configurations in bold (multiple bold configurations indicate lack of a statistically significant difference).}
\label{fig:ged_raters}
\end{figure*}

\section{Results and Discussion}
\label{sec:rnd}
Figure \ref{fig:ged_trends} shows the trend in the GED metric for both datasets. The results show that the optimal choice of model, with respect to the median GED, depends on the dataset. In Figure \ref{fig:ged_lidc}, we see that the models using a mixture of Gaussians to model the latent space distribution (Mixture FC and Mixure FC-LR) perform best with respect to GED for the LIDC-IDRI dataset. In Figure \ref{fig:ged_wmh}, the model ensemble performs best for the WMH dataset.

Furthermore, we see that the different configurations of the Generalized Probabilistic U-Net do considerably better than MC-Dropout/model ensemble for the LIDC-IDRI dataset. For the WMH dataset, we observed very similar performances by all the models studied. The GED results showed that there was no statistically significant difference between any model pair except the model ensemble. 

To explain model performances in Figure \ref{fig:ged_trends}, we studied trends in overlap and diversity terms in the GED equation (Equation \ref{eq:ged}), shown in Figure \ref{fig:ged_breakup}. A model that produces predictions that have a higher overlap capture the common variants, whereas predictions with a higher diversity can capture rarer variants in the output distribution. Therefore, the GED, which essentially captures the \emph{distance} between distributions, is determined by the balance between diversity and overlap.


In Figure \ref{fig:ged_breakup_lidc}, for the LIDC-IDRI dataset, we show that the Generalized Probabilistic U-Net configurations have a lower average overlap, but higher diversity compared with the MC-Dropout, ensemble, and U-Net models. Therefore, these models capture the common outcome well, but fail to produce diverse outputs for the LIDC-IDRI dataset. Our results show that the mixture models (Mixture FC and Mixture FC-LR) achieve an optimal combination of overlap and diversity to emerge as the best performing models with respect to the GED. 

In Figure \ref{fig:ged_breakup_wmh}, there is no statistically significant difference in the overlap trends between a majority of the models. The results show that samples from the model ensemble have a higher sample diversity compared to the rest of the models, leading to better performance with respect to the GED metric.

In Figure \ref{fig:ged_breakup} we see different trends for the LIDC-IDRI and WMH datasets. This relates to the fundamental difference between the datasets, in particular, the structures of interest and the variations therein. The difference between the datasets with respect to the inter-rater agreement can be seen in Figure \ref{fig:image_tiles}. In the example shown, the different raters produce similar segmentations for the WMH dataset, whereas the differences are more noticeable for the LIDC-IDRI example. The inter-rater variability trends can be seen in Figure \ref{fig:ged_breakup}, which show that the WMH dataset has a lower diversity in its labels compared to the LIDC-IDRI datasets. Furthermore, compared to tumors, hyperintensities are smaller structures and therefore any variation present is extremely local. Our results (Figure \ref{fig:ged_trends}) demonstrate that model ensembles and MC-Dropout are better choices to capture spatially local variation, while Probabilistic U-Net variants are better at capturing more long-range variation and increased rater disagreement. 

The consequence of the overlap-diversity balance for the LIDC-IDRI dataset can be seen clearly in Figure \ref{fig:ged_bucket}. For segmentations with larger degree of agreement between raters, that is, where one out of four or four out of four raters agree on the presence of a lesion, the MC-Dropout and ensemble models perform on par with the Generalized Probabilistic U-Net configurations with respect to the GED metric. In the other cases, the Generalized Probabilistic U-Net configurations do better since they are able to better estimate uncertainty by also capturing the rarer samples from the label distribution. This is something the MC-Dropout and ensemble were unable to do, leading to a mismatch between the prediction and label distributions. Intuitively, it can be argued that the latent space of the Generalized Probabilistic U-Net seeks to learn the distribution of labels as part of its training objective, thereby enabling them to model the rarer variants and capture rater disagreement. In contrast, MC-Dropout and model ensembles might produce diverse samples either through sub-sampling of model weights (MC-Dropout) or differing optimization trajectories (model ensembles). This diversity might not correspond to the true distribution of the labels because the objective function for these models usually correspond to maximising accuracy.

Figure \ref{fig:rank_exp} shows the influence of the \emph{rank} hyperparameter on the GED trends for both datasets. Our results show that there is no clear relationship between the rank used to construct the covariance matrix and the trends in the GED metric. A similar observation was reported in \citet{monteiro_stochastic_2020} when the low-rank factor method was used to construct the covariance matrix for the logit (pre-softmax) layer. The results for the LIDC-IDRI dataset show that for all latent space dimensions, there exists an FC-LR configuration that outperforms the FC configuration. The difference between the FC and FC-LR configurations lies in the method used to construct the covariance matrix, which might indicate that the low-rank factor method is a better method. However, the trends for the WMH dataset provide no such indication because there is no statistically significant difference between the FC and best performing FC-LR configurations for the latent space dimensions studied.

Figure \ref{fig:ged_raters} shows changes in the GED metric trends as we increase the number of raters used during training. The trends seen for the Generalized Probabilistic U-Net configurations seem to be intuitive for both datasets, that is, using more than one rater during training leads to an improvement in the GED on the test-set. However, the extent of generalization, that is, the decrease in GED, is different for  the different variants. For MC-Dropout and the model ensemble, we observed little change in the GED trends as the number of raters was increased. This shows that while these methods do produce predictions with some variation, they do not really capture the inter-rater differences in any meaningful manner. 

Our results show that achieving a good match between the prediction and ground truth distributions depends on striking the optimal (with respect to the dataset) balance between overlap and diversity. The structure of interest and variability between raters dictates the choice of the model that achieves that optimal balance. The hyperintensities in the WMH dataset are  uniform in structure, with a larger agreement among raters. The Generalized Probabilistic U-Net configurations fail to produce predictions that match the label distribution, with the model ensemble being the best choice. On the other hand, the tumors in the LIDC-IDRI dataset have a more varied structure (between different patients) and more variability in the segmentations (for the same patient) as compared to the WMH dataset. Our results show that MC-Dropout and the model ensemble fail to capture this variability, and the Generalized Probabilistic U-Net configurations are a better choice. Within the Generalized Probabilistic U-Net configurations, we found models that used a mixture of Gaussians to model the latent space distributions performed best.


All Generalized Probabilistic U-Net configurations showed poor performance in the single rater scenario, when only annotations from one rater were used during training. However, the original Probabilistic U-Net framework was not designed for this scenario. Our results confirm that using the (Generalized) Probabilistic U-Net is a suboptimal choice when segmentations from only one rater are available for training.

We treated the number of Gaussians in the mixture model as a hyperparameter that was fixed at the start of training. Future implementations could treat it as a learned parameter, where the final number of Gaussians can be optimized during training. Similar to \citet{kohl_probabilistic_2018} and \citet{baumgartner_phiseg_2019}, we used random sampling to select ground truth segmentations during training. Studying the interplay between different label fusion methods~\citep{warfield_simultaneous_2004, shen_improving_2019} and latent space distributions is an interesting research direction. Our extensions to model more general forms of the Gaussian distribution can easily be integrated into a hierarchical latent space model such as PHISeg~\citep{baumgartner_phiseg_2019}. Future research could be directed at studying the efficacy of different latent space distributions at different resolutions in a hierarchical model.

Our results show that the Generalized Probabilistic U-Net is a good choice for segmenting ambiguous medical images, where ambiguity is quantified by disagreement between annotators. Furthermore, choosing more general forms of the Gaussian distribution as the latent space distribution results in statistically significant improvements over the default axis-aligned distribution, as measured by the GED metric on the (more ambiguous) LIDC-IDRI dataset. It is therefore advised to explore alternative Gaussian distributions, since the default choice of axis-aligned Gaussian for the latent space distribution is never the optimal choice in our experiments.

\section{Conclusions}
Our work focused on extending the Probabilistic U-Net framework by going beyond the axis-aligned Gaussian distribution as the de facto choice for the variational distribution. Our results showed that the choice of model (or latent space distribution) is dictated by the optimal overlap-diversity combination for a particular dataset. Therefore, investigating the suitability of different latent space distributions for a particular dataset is beneficial. The distributions studied in this paper can be used as a drop-in replacement for the axis-aligned Gaussian in other extensions to the Probabilistic U-Net framework, like the class of models using hierarchical latent spaces~\citep{baumgartner_phiseg_2019}, or as base distributions for normalizing flow-based segmentation models~\citep{liu_uncertainty_2020}.

\section{Acknowledgements}
This work was financially supported by the project IMPACT (Intelligence based iMprovement of Personalized treatment And Clinical workflow supporT) in the framework of the EU research programme ITEA3 (Information Technology for European Advancement).

\ethics{The work follows appropriate ethical standards in conducting research and writing the manuscript, following all applicable laws and regulations regarding treatment of animals or human subjects. The UMCU Medical Ethical Committee has reviewed this study and informed consent was waived due to its retrospective nature.}

\coi{The authors declare no conflict of interest.}

\bibliography{sample}

\begin{thebibliography}{46}
\providecommand{\natexlab}[1]{#1}
\providecommand{\url}[1]{\texttt{#1}}
\expandafter\ifx\csname urlstyle\endcsname\relax
  \providecommand{\doi}[1]{doi: #1}\else
  \providecommand{\doi}{doi: \begingroup \urlstyle{rm}\Url}\fi

\bibitem[Armato et~al.(2011)Armato, McLennan, Bidaut, McNitt-Gray, Meyer,
  Reeves, Zhao, Aberle, et~al.]{lidc2011}
Samuel~G. Armato, Geoffrey McLennan, Luc Bidaut, Michael~F. McNitt-Gray,
  Charles~R. Meyer, Anthony~P. Reeves, Binsheng Zhao, Denise~R. Aberle, et~al.
\newblock The {Lung} {Image} {Database} {Consortium} ({LIDC}) and {Image}
  {Database} {Resource} {Initiative} ({IDRI}): {A} {Completed} {Reference}
  {Database} of {Lung} {Nodules} on {CT} {Scans}.
\newblock \emph{Medical Physics}, 38\penalty0 (2):\penalty0 915--931, February
  2011.
\newblock ISSN 0094-2405.
\newblock \doi{10.1118/1.3528204}.

\bibitem[{Armen Der Kiureghian} and
  Ditlevsen(2009)]{armen_der_kiureghian_aleatory_2009}
{Armen Der Kiureghian} and Ove Ditlevsen.
\newblock Aleatory or epistemic? {Does} it matter?
\newblock \emph{Structural Safety}, 31\penalty0 (2):\penalty0 105--112, March
  2009.
\newblock ISSN 0167-4730.
\newblock \doi{10.1016/j.strusafe.2008.06.020}.

\bibitem[Barber and Bishop(1997)]{barber_ensemble_1998}
David Barber and Christopher Bishop.
\newblock Ensemble learning for multi-layer networks.
\newblock In M.~Jordan, M.~Kearns, and S.~Solla, editors, \emph{Advances in
  Neural Information Processing Systems}, volume~10, pages 395 -- 401. MIT
  Press, 1997.
\newblock URL
  \url{https://proceedings.neurips.cc/paper/1997/file/e816c635cad85a60fabd6b97b03cbcc9-Paper.pdf}.

\bibitem[Baumgartner et~al.(2019)Baumgartner, Tezcan, Chaitanya, Hötker,
  Muehlematter, Schawkat, Becker, Donati, and
  Konukoglu]{baumgartner_phiseg_2019}
Christian~F. Baumgartner, Kerem~C. Tezcan, Krishna Chaitanya, Andreas~M.
  Hötker, Urs~J. Muehlematter, Khoschy Schawkat, Anton~S. Becker, Olivio
  Donati, and Ender Konukoglu.
\newblock {PHiSeg}: {Capturing} {Uncertainty} in {Medical} {Image}
  {Segmentation}.
\newblock In \emph{Medical {Image} {Computing} and {Computer} {Assisted}
  {Intervention} – {MICCAI} 2019}, Lecture {Notes} in {Computer} {Science},
  pages 119--127. Springer International Publishing, 2019.
\newblock ISBN 978-3-030-32245-8.
\newblock \doi{10.1007/978-3-030-32245-8_14}.

\bibitem[Bhat et~al.(2022)Bhat, Pluim, and Kuijf]{bhat_generalized_2022}
Ishaan Bhat, Josien P.~W. Pluim, and Hugo~J. Kuijf.
\newblock Generalized probabilistic u-net for medical image segementation.
\newblock In Carole~H. Sudre, Christian~F. Baumgartner, Adrian Dalca, Chen Qin,
  Ryutaro Tanno, Koen Van~Leemput, and William~M. Wells~III, editors,
  \emph{Uncertainty for Safe Utilization of Machine Learning in Medical
  Imaging}, pages 113--124, Cham, 2022. Springer Nature Switzerland.
\newblock ISBN 978-3-031-16749-2.
\newblock \doi{10.1007/978-3-031-16749-2_11}.

\bibitem[Bishop(2006)]{bishop2006}
Christopher~M. Bishop.
\newblock \emph{Pattern Recognition and Machine Learning (Information Science
  and Statistics)}.
\newblock Springer-Verlag, 2006.
\newblock ISBN:978-0-387-31073-2.

\bibitem[Blundell et~al.(2015)Blundell, Cornebise, Kavukcuoglu, and
  Wierstra]{blundell_weight_2015}
Charles Blundell, Julien Cornebise, Koray Kavukcuoglu, and Daan Wierstra.
\newblock Weight uncertainty in neural network.
\newblock In Francis Bach and David Blei, editors, \emph{Proceedings of the
  32nd International Conference on Machine Learning}, volume~37, pages
  1613--1622. PMLR, 2015.
\newblock URL \url{https://proceedings.mlr.press/v37/blundell15.html}.

\bibitem[Chotzoglou and Kainz(2019)]{zhou_exploring_2019}
Elisa Chotzoglou and Bernhard Kainz.
\newblock Exploring the {Relationship} {Between} {Segmentation} {Uncertainty},
  {Segmentation} {Performance} and {Inter}-observer {Variability} with
  {Probabilistic} {Networks}.
\newblock In \emph{Large-{Scale} {Annotation} of {Biomedical} {Data} and
  {Expert} {Label} {Synthesis} and {Hardware} {Aware} {Learning} for {Medical}
  {Imaging} and {Computer} {Assisted} {Intervention}}, volume 11851, pages
  51--60. Springer International Publishing, 2019.
\newblock \doi{10.1007/978-3-030-33642-4_6}.

\bibitem[Gal and Ghahramani(2016)]{galmcd}
Yarin Gal and Zoubin Ghahramani.
\newblock Dropout as a bayesian approximation: Representing model uncertainty
  in deep learning.
\newblock In Maria~Florina Balcan and Kilian~Q. Weinberger, editors,
  \emph{Proceedings of The 33rd International Conference on Machine Learning},
  volume~48 of \emph{Proceedings of Machine Learning Research}, pages
  1050--1059. PMLR, 20--22 Jun 2016.
\newblock URL \url{https://proceedings.mlr.press/v48/gal16.html}.

\bibitem[Hershey and Olsen(2007)]{hershey2007}
John~R. Hershey and Peder~A. Olsen.
\newblock Approximating the kullback leibler divergence between gaussian
  mixture models.
\newblock In \emph{2007 IEEE International Conference on Acoustics, Speech and
  Signal Processing - ICASSP '07}, volume~4, pages IV--317--IV--320, 2007.
\newblock \doi{10.1109/ICASSP.2007.366913}.

\bibitem[Higgins et~al.(2017)Higgins, Matthey, Pal, Burgess, Glorot, Botvinick,
  Mohamed, and Lerchner]{Higgins2017betaVAELB}
Irina Higgins, Loic Matthey, Arka Pal, Christopher Burgess, Xavier Glorot,
  Matthew Botvinick, Shakir Mohamed, and Alexander Lerchner.
\newblock beta-{VAE}: Learning basic visual concepts with a constrained
  variational framework.
\newblock In \emph{International Conference on Learning Representations}, 2017.
\newblock URL \url{https://openreview.net/forum?id=Sy2fzU9gl}.

\bibitem[Hu et~al.(2019)Hu, Worrall, Knegt, Veeling, Huisman, and
  Welling]{hu_supervised_2019}
Shi Hu, Daniel Worrall, Stefan Knegt, Bas Veeling, Henkjan Huisman, and Max
  Welling.
\newblock Supervised {Uncertainty} {Quantification} for {Segmentation} with
  {Multiple} {Annotations}.
\newblock In \emph{Medical {Image} {Computing} and {Computer} {Assisted}
  {Intervention} – {MICCAI} 2019}, Lecture {Notes} in {Computer} {Science},
  pages 137--145. Springer International Publishing, 2019.
\newblock ISBN 978-3-030-32245-8.
\newblock \doi{10.1007/978-3-030-32245-8_16}.

\bibitem[Iglesias et~al.(2011)Iglesias, Liu, Thompson, and
  Tu]{iglesias2011robust}
Juan~Eugenio Iglesias, Cheng-Yi Liu, Paul~M. Thompson, and Zhuowen Tu.
\newblock Robust brain extraction across datasets and comparison with publicly
  available methods.
\newblock \emph{IEEE Transactions on Medical Imaging}, 30\penalty0
  (9):\penalty0 1617--1634, 2011.
\newblock \doi{10.1109/TMI.2011.2138152}.

\bibitem[Ioffe and Szegedy(2015)]{ioffe15}
Sergey Ioffe and Christian Szegedy.
\newblock Batch normalization: Accelerating deep network training by reducing
  internal covariate shift.
\newblock In Francis Bach and David Blei, editors, \emph{Proceedings of the
  32nd International Conference on Machine Learning}, volume~37 of
  \emph{Proceedings of Machine Learning Research}, pages 448--456. PMLR, 2015.
\newblock URL \url{https://proceedings.mlr.press/v37/ioffe15.html}.

\bibitem[Jang et~al.(2017)Jang, Gu, and Poole]{jang_categorical_2017}
Eric Jang, Shixiang Gu, and Ben Poole.
\newblock Categorical reparameterization with gumbel-softmax.
\newblock In \emph{International Conference on Learning Representations}, 2017.
\newblock URL \url{https://openreview.net/forum?id=rkE3y85ee}.

\bibitem[Jensen et~al.(2019)Jensen, Jørgensen, Jalaboi, Hansen, and
  Olsen]{shen_improving_2019}
Martin~Holm Jensen, Dan~Richter Jørgensen, Raluca Jalaboi, Mads~Eiler Hansen,
  and Martin~Aastrup Olsen.
\newblock Improving {Uncertainty} {Estimation} in {Convolutional} {Neural}
  {Networks} {Using} {Inter}-rater {Agreement}.
\newblock In \emph{Medical {Image} {Computing} and {Computer} {Assisted}
  {Intervention} – {MICCAI} 2019}, volume 11767, pages 540--548. Springer
  International Publishing, 2019.
\newblock \doi{10.1007/978-3-030-32251-9_59}.

\bibitem[Jiang et~al.(2017)Jiang, Zheng, Tan, Tang, and
  Zhou]{jiang_variational_2017}
Zhuxi Jiang, Yin Zheng, Huachun Tan, Bangsheng Tang, and Hanning Zhou.
\newblock Variational {Deep} {Embedding}: {An} {Unsupervised} and {Generative}
  {Approach} to {Clustering}.
\newblock In \emph{Proceedings of the {Twenty}-{Sixth} {International} {Joint}
  {Conference} on {Artificial} {Intelligence}}, pages 1965--1972. International
  Joint Conferences on Artificial Intelligence Organization, August 2017.
\newblock ISBN 978-0-9992411-0-3.
\newblock \doi{10.24963/ijcai.2017/273}.

\bibitem[Jungo et~al.(2018)Jungo, Meier, Ermis, Blatti-Moreno, Herrmann, Wiest,
  and Reyes]{jungo_effect_2018}
Alain Jungo, Raphael Meier, Ekin Ermis, Marcela Blatti-Moreno, Evelyn Herrmann,
  Roland Wiest, and Mauricio Reyes.
\newblock On the {Effect} of {Inter}-observer {Variability} for a {Reliable}
  {Estimation} of {Uncertainty} of {Medical} {Image} {Segmentation}.
\newblock In \emph{Medical {Image} {Computing} and {Computer} {Assisted}
  {Intervention} – {MICCAI} 2018}, pages 682--690. Springer International
  Publishing, 2018.
\newblock \doi{10.1007/978-3-030-00928-1_77}.

\bibitem[Jungo et~al.(2020)Jungo, Balsiger, and Reyes]{jungo_analyzing_2020}
Alain Jungo, Fabian Balsiger, and Mauricio Reyes.
\newblock Analyzing the {Quality} and {Challenges} of {Uncertainty}
  {Estimations} for {Brain} {Tumor} {Segmentation}.
\newblock \emph{Frontiers in Neuroscience}, 14, April 2020.
\newblock ISSN 1662-453X.
\newblock \doi{10.3389/fnins.2020.00282}.

\bibitem[Kamnitsas et~al.(2018)Kamnitsas, Bai, Ferrante, McDonagh, Sinclair,
  Pawlowski, Rajchl, Lee, Kainz, Rueckert, and
  Glocker]{kamnitsas_ensembles_2018}
K.~Kamnitsas, W.~Bai, E.~Ferrante, S.~McDonagh, M.~Sinclair, N.~Pawlowski,
  M.~Rajchl, M.~Lee, B.~Kainz, D.~Rueckert, and B.~Glocker.
\newblock Ensembles of multiple models and architectures for robust brain
  tumour segmentation.
\newblock In Alessandro Crimi, Spyridon Bakas, Hugo Kuijf, Bjoern Menze, and
  Mauricio Reyes, editors, \emph{Brainlesion: Glioma, Multiple Sclerosis,
  Stroke and Traumatic Brain Injuries}, pages 450--462, Cham, 2018. Springer
  International Publishing.
\newblock ISBN 978-3-319-75238-9.
\newblock \doi{10.1007/978-3-319-75238-9_38}.

\bibitem[Kendall and Gal(2017)]{kendall_what_2017}
Alex Kendall and Yarin Gal.
\newblock What uncertainties do we need in bayesian deep learning for computer
  vision?
\newblock In I.~Guyon, U.~Von Luxburg, S.~Bengio, H.~Wallach, R.~Fergus,
  S.~Vishwanathan, and R.~Garnett, editors, \emph{Advances in Neural
  Information Processing Systems}, volume~30. Curran Associates, Inc., 2017.
\newblock URL
  \url{https://proceedings.neurips.cc/paper/2017/file/2650d6089a6d640c5e85b2b88265dc2b-Paper.pdf}.

\bibitem[Kingma and Ba(2015)]{kingma_adam:_2015}
Diederik~P. Kingma and Jimmy Ba.
\newblock Adam: {A} {Method} for {Stochastic} {Optimization}.
\newblock In \emph{International Conference on Learning Representations}, 2015.
\newblock URL \url{https://hdl.handle.net/11245/1.505367}.

\bibitem[Kingma and Welling(2014)]{Kingma2014}
Diederik~P. Kingma and Max Welling.
\newblock {Auto-Encoding Variational Bayes}.
\newblock In \emph{International Conference on Learning Representations}, 2014.
\newblock URL \url{https://hdl.handle.net/11245/1.434281}.

\bibitem[Kingma and Welling(2019)]{kingma_introduction_2019}
Diederik~P. Kingma and Max Welling.
\newblock An {Introduction} to {Variational} {Autoencoders}.
\newblock \emph{Foundations and Trends® in Machine Learning}, 12\penalty0
  (4):\penalty0 307--392, 2019.
\newblock ISSN 1935-8237, 1935-8245.
\newblock \doi{10.1561/2200000056}.

\bibitem[Kingma et~al.(2015)Kingma, Salimans, and
  Welling]{kingma_variational_2015}
Durk~P Kingma, Tim Salimans, and Max Welling.
\newblock Variational {Dropout} and the {Local} {Reparameterization} {Trick}.
\newblock In C.~Cortes, N.~Lawrence, D.~Lee, M.~Sugiyama, and R.~Garnett,
  editors, \emph{Advances in {Neural} {Information} {Processing} {Systems}},
  volume~28. Curran Associates, Inc., 2015.
\newblock URL
  \url{https://proceedings.neurips.cc/paper/2015/hash/bc7316929fe1545bf0b98d114ee3ecb8-Abstract.html}.

\bibitem[Klein et~al.(2010)Klein, Staring, Murphy, Viergever, and
  Pluim]{elastix2010}
S.~Klein, M.~Staring, K.~Murphy, M.A. Viergever, and J.P.W. Pluim.
\newblock Elastix : a toolbox for intensity-based medical image registration.
\newblock \emph{IEEE Transactions on Medical Imaging}, 29\penalty0
  (1):\penalty0 196--205, 2010.
\newblock ISSN 0278-0062.
\newblock \doi{10.1109/TMI.2009.2035616}.

\bibitem[Kohl et~al.(2018)Kohl, Romera-Paredes, Meyer, De~Fauw, Ledsam,
  Maier-Hein, Eslami, Jimenez~Rezende, and
  Ronneberger]{kohl_probabilistic_2018}
Simon Kohl, Bernardino Romera-Paredes, Clemens Meyer, Jeffrey De~Fauw,
  Joseph~R. Ledsam, Klaus Maier-Hein, S.~M.~Ali Eslami, Danilo Jimenez~Rezende,
  and Olaf Ronneberger.
\newblock A {Probabilistic} {U}-{Net} for {Segmentation} of {Ambiguous}
  {Images}.
\newblock In \emph{Advances in {Neural} {Information} {Processing} {Systems}},
  volume~31. Curran Associates, Inc., 2018.
\newblock URL
  \url{https://papers.nips.cc/paper/2018/hash/473447ac58e1cd7e96172575f48dca3b-Abstract.html}.

\bibitem[Kopf et~al.(2021)Kopf, Fortuin, Somnath, and
  Claassen]{kopf_mixture--experts_2021}
Andreas Kopf, Vincent Fortuin, Vignesh~Ram Somnath, and Manfred Claassen.
\newblock Mixture-of-{Experts} {Variational} {Autoencoder} for clustering and
  generating from similarity-based representations on single cell data.
\newblock \emph{PLOS Computational Biology}, 17\penalty0 (6):\penalty0
  e1009086, June 2021.
\newblock ISSN 1553-7358.
\newblock \doi{10.1371/journal.pcbi.1009086}.

\bibitem[Kuijf et~al.(2019)Kuijf, Biesbroek, De~Bresser, Heinen, Andermatt,
  Bento, Berseth, Belyaev, Cardoso, Casamitjana, Collins, Dadar, Georgiou,
  Ghafoorian, Jin, Khademi, Knight, Li, Lladó, Luna, Mahmood, McKinley,
  Mehrtash, Ourselin, Park, Park, Park, Pezold, Puybareau, Rittner, Sudre,
  Valverde, Vilaplana, Wiest, Xu, Xu, Zeng, Zhang, Zheng, Chen, van~der Flier,
  Barkhof, Viergever, and Biessels]{kuijf2019}
Hugo~J. Kuijf, J.~Matthijs Biesbroek, Jeroen De~Bresser, Rutger Heinen, Simon
  Andermatt, Mariana Bento, Matt Berseth, Mikhail Belyaev, M.~Jorge Cardoso,
  Adrià Casamitjana, D.~Louis Collins, Mahsa Dadar, Achilleas Georgiou, Mohsen
  Ghafoorian, Dakai Jin, April Khademi, Jesse Knight, Hongwei Li, Xavier
  Lladó, Miguel Luna, Qaiser Mahmood, Richard McKinley, Alireza Mehrtash,
  Sébastien Ourselin, Bo-Yong Park, Hyunjin Park, Sang~Hyun Park, Simon
  Pezold, Elodie Puybareau, Leticia Rittner, Carole~H. Sudre, Sergi Valverde,
  Verónica Vilaplana, Roland Wiest, Yongchao Xu, Ziyue Xu, Guodong Zeng,
  Jianguo Zhang, Guoyan Zheng, Christopher Chen, Wiesje van~der Flier, Frederik
  Barkhof, Max~A. Viergever, and Geert~Jan Biessels.
\newblock Standardized assessment of automatic segmentation of white matter
  hyperintensities and results of the wmh segmentation challenge.
\newblock \emph{IEEE Transactions on Medical Imaging}, 38\penalty0
  (11):\penalty0 2556--2568, 2019.
\newblock \doi{10.1109/TMI.2019.2905770}.

\bibitem[Lakshminarayanan et~al.(2017)Lakshminarayanan, Pritzel, and
  Blundell]{lakshminarayanan_simple_2017}
Balaji Lakshminarayanan, Alexander Pritzel, and Charles Blundell.
\newblock Simple and {Scalable} {Predictive} {Uncertainty} {Estimation} using
  {Deep} {Ensembles}.
\newblock In \emph{Advances in {Neural} {Information} {Processing} {Systems}},
  volume~30. Curran Associates, Inc., 2017.
\newblock URL
  \url{https://proceedings.neurips.cc/paper/2017/file/9ef2ed4b7fd2c810847ffa5fa85bce38-Paper.pdf}.

\bibitem[Maddison et~al.(2017)Maddison, Mnih, and Teh]{maddison_concrete_2017}
Chris~J. Maddison, Andriy Mnih, and Yee~Whye Teh.
\newblock The concrete distribution: A continuous relaxation of discrete random
  variables.
\newblock In \emph{International Conference on Learning Representations}, 2017.
\newblock URL \url{https://openreview.net/forum?id=S1jE5L5gl}.

\bibitem[Mehrtash et~al.(2020)Mehrtash, Wells, Tempany, Abolmaesumi, and
  Kapur]{mehrtash2020}
Alireza Mehrtash, William~M. Wells, Clare~M. Tempany, Purang Abolmaesumi, and
  Tina Kapur.
\newblock Confidence calibration and predictive uncertainty estimation for deep
  medical image segmentation.
\newblock \emph{IEEE Transactions on Medical Imaging}, 39\penalty0
  (12):\penalty0 3868--3878, 2020.
\newblock \doi{10.1109/TMI.2020.3006437}.

\bibitem[Monteiro et~al.(2020)Monteiro, Le~Folgoc, Coelho~de Castro, Pawlowski,
  Marques, Kamnitsas, van~der Wilk, and Glocker]{monteiro_stochastic_2020}
Miguel Monteiro, Loic Le~Folgoc, Daniel Coelho~de Castro, Nick Pawlowski,
  Bernardo Marques, Konstantinos Kamnitsas, Mark van~der Wilk, and Ben Glocker.
\newblock Stochastic segmentation networks: Modelling spatially correlated
  aleatoric uncertainty.
\newblock In H.~Larochelle, M.~Ranzato, R.~Hadsell, M.F. Balcan, and H.~Lin,
  editors, \emph{Advances in Neural Information Processing Systems}, volume~33,
  pages 12756--12767. Curran Associates, Inc., 2020.
\newblock URL
  \url{https://proceedings.neurips.cc/paper/2020/file/95f8d9901ca8878e291552f001f67692-Paper.pdf}.

\bibitem[Paszke et~al.(2019)Paszke, Gross, Massa, Lerer, Bradbury, Chanan,
  Killeen, Lin, Gimelshein, Antiga, Desmaison, Kopf, Yang, DeVito, Raison,
  Tejani, Chilamkurthy, Steiner, Fang, Bai, and Chintala]{pytorch2019}
Adam Paszke, Sam Gross, Francisco Massa, Adam Lerer, James Bradbury, Gregory
  Chanan, Trevor Killeen, Zeming Lin, Natalia Gimelshein, Luca Antiga, Alban
  Desmaison, Andreas Kopf, Edward Yang, Zachary DeVito, Martin Raison, Alykhan
  Tejani, Sasank Chilamkurthy, Benoit Steiner, Lu~Fang, Junjie Bai, and Soumith
  Chintala.
\newblock Pytorch: An imperative style, high-performance deep learning library.
\newblock In H.~Wallach, H.~Larochelle, A.~Beygelzimer, F.~d\textquotesingle
  Alch\'{e}-Buc, E.~Fox, and R.~Garnett, editors, \emph{Advances in Neural
  Information Processing Systems 32}, pages 8024--8035. Curran Associates,
  Inc., 2019.
\newblock URL
  \url{https://papers.nips.cc/paper/2019/hash/bdbca288fee7f92f2bfa9f7012727740-Abstract.html}.

\bibitem[Rezende and Mohamed(2015)]{rezende_variational_2016}
Danilo Rezende and Shakir Mohamed.
\newblock Variational inference with normalizing flows.
\newblock In Francis Bach and David Blei, editors, \emph{Proceedings of the
  32nd International Conference on Machine Learning}, volume~37 of
  \emph{Proceedings of Machine Learning Research}, pages 1530--1538. PMLR,
  2015.
\newblock URL \url{https://proceedings.mlr.press/v37/rezende15.html}.

\bibitem[Rezende et~al.(2014)Rezende, Mohamed, and
  Wierstra]{pmlr-v32-rezende14}
Danilo~Jimenez Rezende, Shakir Mohamed, and Daan Wierstra.
\newblock Stochastic backpropagation and approximate inference in deep
  generative models.
\newblock In Eric~P. Xing and Tony Jebara, editors, \emph{Proceedings of the
  31st International Conference on Machine Learning}, volume~32 of
  \emph{Proceedings of Machine Learning Research}, pages 1278--1286. PMLR,
  2014.
\newblock URL \url{https://proceedings.mlr.press/v32/rezende14.html}.

\bibitem[Ronneberger et~al.(2015)Ronneberger, Fischer, and
  Brox]{ronneberger_u-net_2015}
Olaf Ronneberger, Philipp Fischer, and Thomas Brox.
\newblock U-{Net}: {Convolutional} {Networks} for {Biomedical} {Image}
  {Segmentation}.
\newblock In \emph{Medical {Image} {Computing} and {Computer}-{Assisted}
  {Intervention} – {MICCAI} 2015}, pages 234--241. Springer International
  Publishing, 2015.
\newblock ISBN 978-3-319-24574-4.
\newblock \doi{10.1007/978-3-319-24574-4_28}.

\bibitem[Rupprecht et~al.(2017)Rupprecht, Laina, DiPietro, Baust, Tombari,
  Navab, and Hager]{rupprecht_learning_2017}
Christian Rupprecht, Iro Laina, Robert DiPietro, Maximilian Baust, Federico
  Tombari, Nassir Navab, and Gregory~D. Hager.
\newblock Learning in an {Uncertain} {World}: {Representing} {Ambiguity}
  {Through} {Multiple} {Hypotheses}.
\newblock In \emph{2017 {IEEE} {International} {Conference} on {Computer}
  {Vision} ({ICCV})}, pages 3611--3620, October 2017.
\newblock \doi{10.1109/ICCV.2017.388}.

\bibitem[Selvan et~al.(2020)Selvan, Faye, Middleton, and
  Pai]{liu_uncertainty_2020}
Raghavendra Selvan, Frederik Faye, Jon Middleton, and Akshay Pai.
\newblock Uncertainty quantification in medical image segmentation with
  normalizing flows.
\newblock In Mingxia Liu, Pingkun Yan, Chunfeng Lian, and Xiaohuan Cao,
  editors, \emph{Machine Learning in Medical Imaging}, pages 80--90, Cham,
  2020. Springer International Publishing.
\newblock ISBN 978-3-030-59861-7.
\newblock \doi{10.1007/978-3-030-59861-7_9}.

\bibitem[Sohn et~al.(2015)Sohn, Lee, and Yan]{sohn_learning_2015}
Kihyuk Sohn, Honglak Lee, and Xinchen Yan.
\newblock Learning {Structured} {Output} {Representation} using {Deep}
  {Conditional} {Generative} {Models}.
\newblock In \emph{Advances in {Neural} {Information} {Processing} {Systems}},
  volume~28. Curran Associates, Inc., 2015.
\newblock URL
  \url{https://proceedings.neurips.cc/paper/2015/file/8d55a249e6baa5c06772297520da2051-Paper.pdf}.

\bibitem[Srivastava et~al.(2014)Srivastava, Hinton, Krizhevsky, Sutskever, and
  Salakhutdinov]{srivastava_dropout_2014}
Nitish Srivastava, Geoffrey Hinton, Alex Krizhevsky, Ilya Sutskever, and Ruslan
  Salakhutdinov.
\newblock Dropout: A simple way to prevent neural networks from overfitting.
\newblock \emph{Journal of Machine Learning Research}, 15\penalty0
  (56):\penalty0 1929--1958, 2014.
\newblock URL \url{http://jmlr.org/papers/v15/srivastava14a.html}.

\bibitem[Teye et~al.(2018)Teye, Azizpour, and Smith]{pmlr-v80-teye18a}
Mattias Teye, Hossein Azizpour, and Kevin Smith.
\newblock {B}ayesian uncertainty estimation for batch normalized deep networks.
\newblock In Jennifer Dy and Andreas Krause, editors, \emph{Proceedings of the
  35th International Conference on Machine Learning}, volume~80 of
  \emph{Proceedings of Machine Learning Research}, pages 4907--4916. PMLR,
  2018.
\newblock URL \url{https://proceedings.mlr.press/v80/teye18a.html}.

\bibitem[Wainwright and Jordan(2007)]{wainwright_graphical_2007}
Martin~J. Wainwright and Michael~I. Jordan.
\newblock Graphical {Models}, {Exponential} {Families}, and {Variational}
  {Inference}.
\newblock \emph{Foundations and Trends® in Machine Learning}, 1\penalty0
  (1–2):\penalty0 1--305, 2007.
\newblock ISSN 1935-8237, 1935-8245.
\newblock \doi{10.1561/2200000001}.

\bibitem[Wang et~al.(2019)Wang, Li, Aertsen, Deprest, Ourselin, and
  Vercauteren]{wang_aleatoric_2019}
Guotai Wang, Wenqi Li, Michael Aertsen, Jan Deprest, Sébastien Ourselin, and
  Tom Vercauteren.
\newblock Aleatoric uncertainty estimation with test-time augmentation for
  medical image segmentation with convolutional neural networks.
\newblock \emph{Neurocomputing}, 338:\penalty0 34 -- 45, 2019.
\newblock ISSN 0925-2312.
\newblock \doi{https://doi.org/10.1016/j.neucom.2019.01.103}.

\bibitem[Warfield et~al.(2004)Warfield, Zou, and
  Wells]{warfield_simultaneous_2004}
S.K. Warfield, K.H. Zou, and W.M. Wells.
\newblock Simultaneous truth and performance level estimation ({STAPLE}): an
  algorithm for the validation of image segmentation.
\newblock \emph{IEEE Transactions on Medical Imaging}, 23\penalty0
  (7):\penalty0 903--921, July 2004.
\newblock ISSN 1558-254X.
\newblock \doi{10.1109/TMI.2004.828354}.

\bibitem[Williams(1996)]{williams1996}
Peter~M. Williams.
\newblock Using neural networks to model conditional multivariate densities.
\newblock \emph{Neural Comput.}, 8\penalty0 (4):\penalty0 843–854, may 1996.
\newblock ISSN 0899-7667.
\newblock \doi{10.1162/neco.1996.8.4.843}.

\end{thebibliography}
\end{document}